\documentclass{article} 
\usepackage{iclr2025_re-align_workshop,times}

\usepackage{amsmath,amsfonts,bm}

\def\eqref#1{equation~\ref{#1}}

\def\1{\bm{1}}

\DeclareMathAlphabet{\mathsfit}{\encodingdefault}{\sfdefault}{m}{sl}
\SetMathAlphabet{\mathsfit}{bold}{\encodingdefault}{\sfdefault}{bx}{n}

\usepackage{hyperref}
\usepackage{url}
\usepackage{amsmath}
\usepackage{amsfonts}
\usepackage{float}
\usepackage{graphicx}
\usepackage{lipsum}
\usepackage{accents}
\usepackage{xspace}

\title{The Spotlight Resonance Method:\\ Resolving the Alignment of Embedded Activations}

\author{George Bird\\
Department of Computer Science\\
University of Manchester\\
Manchester, UK \\
\texttt{george.bird@postgrad.manchester.ac.uk} \\
}

\usepackage{xspace}
\usepackage[colorinlistoftodos, color=blue!30!white, 
]{todonotes}                                        

\iclrfinalcopy 
\begin{document}

\maketitle
\begin{abstract}
    Understanding how deep learning models represent data is currently difficult due to the limited number of methodologies available.
    This paper demonstrates a versatile and novel visualisation tool for determining the axis alignment of embedded data at any layer in any deep learning model. 
    In particular, it evaluates the distribution around planes defined by the network's privileged basis vectors.
    This method provides both an atomistic and a holistic, intuitive metric for interpreting the distribution of activations across all planes. It ensures that both positive and negative signals contribute, treating the activation vector as a whole. Depending on the application, several variations of this technique are presented, with a resolution scale hyperparameter to probe different angular scales. 
    Using this method, multiple examples are provided that demonstrate embedded representations tend to be axis-aligned with the privileged basis. This is not necessarily the standard basis, and it is found that activation functions directly result in privileged bases. Hence, it provides a direct causal link between functional form symmetry breaking and representational alignment, explaining why representations have a tendency to align with the neuron basis. Therefore, using this method, we begin to answer the fundamental question of what causes the observed tendency of representations to align with neurons. Finally, examples of so-called grandmother neurons are found in a variety of networks.
\end{abstract}

\section{Introduction}
    This work aims to better understand how artificial neural networks represent human-interpretable concepts embedded in their hidden layers. Introductory texts often state that individual artificial neurons may respond to distinct real-world signals. This may be a visual neuron that responds to the presence of fur, while another responds to grass. This has been termed a neural ``local coding scheme'' \citep{Foldiak2008}, ``grandmother neurons'' \citep{Gross2002, Connor2005}, ``gnostic neurons'' \citep{Konorski1968} and sometimes ``one-hot encoding'' --- depending on the research field. It is unclear whether trained artificial neural networks produce this structure or whether this is an oversimplification. This work provides a versatile new tool and evidence to aid in determining this fundamental question. 
    
    Samples provided to a neural network are represented as vectors of activations. These are then typically transformed through a series of affine and non-linear steps to achieve the desired result of training. The activation vectors are frequently decomposed into a particular basis for applying the non-linearities. This basis is typically the standard (Kronecker) basis. Each unit vector of the standard basis, $\hat{e}_i$, is typically defined as an individual neuron whose response is often suggested to represent a human-interpretable concept. The standard basis is a common instance of this more general concept of a privileged basis. The term `privileged' is generalised from \citet{Elhage2022}'s paper: it indicates that some model property incurs a unique, inherent basis, which may significantly predispose the model to a particular arrangement of its embeddings across all samples. This privileged basis is a collection of directions where an activation function (or other function) has caused anisotropy around them; this makes these directions unique and stand out to the network. Therefore, the network may alter its embedded distributions in response. For example, elementwise Tanh (and most activation functions) would privilege this standard basis due to the elementwise application; consequently, activations may cluster along the standard basis. This may result in neurons often being associated with specific activations and concepts, reinforcing the observation for representations tending to align with the standard basis. This, in turn, supports the continued use of the standard basis in decompositions and current functional forms. The tool presented detects whether embedded activations preferentially cluster around vectors of the privileged basis after training. This work confirms the above hypothesis that functional form choices privilege a particular basis, which explains why it might be expected that the standard decomposition is typically special due to elementwise application privileging this basis. This conclusion is achieved by inducing a non-standard privileged basis through new activation functions, with distributions observed to cluster only around this new basis. This demonstrates that the standard basis appears special solely because of functional form privileging. Hence, the privileged basis is more fundamental and predictive for representational alignment.
    
    A privileging of the standard basis is expected due to the elementwise nature of current activation functions. This is because the (non-polynomial) non-linearity of the activation functions is essential in approximating arbitrary functions and hence performing the desired computation. If an elementwise activation function is used, anisotropies result around the standard basis vectors, breaking the space's rotational symmetry. Since the desired computation is typically achieved and, therefore, dependent on the use of these functions, it may be expected that anisotropy in the distribution of embedded activations will also be induced during training by such functions --- producing observations of representational alignment. This is expected to then cause a (detectable) increase in density for a (sub)set of embedded activations about these anisotropies, which can suggest a local coding scheme for the network layer. Therefore, if the activations depend on the privileged basis, they may align or anti-align with this basis (the extrema). Alternatively, if independent, they may appear uniformly distributed or uncorrelated with the privileged basis. This phenomenon can be directly measured using the methods proposed in this paper and, therefore, can be used to determine whether neurons correspond to particular human-interpretable concepts across any model. Moreover, this will also be shown to generalise for more cases than just the standard basis. 
    
    This question has been explored numerous times before with as many methods. Some authors find neurons do represent single ideas \citep{Zhou2015}, some authors find no alignment \citep{Szegedy2014}, and sometimes differing arrangements are observed \citep{Papyan2020, Elhage2022, Kothapalli2023}. Yet, on the whole, there is an emerging consensus that there is at least some tendency of neural networks to produce a local coding representation \citep{Vondrick2016, Bau2017, Olah2019, Elhage2022}. Therefore, this question is far from concluded and requires further methodologies for new evidence. Presented in this paper: ``The Spotlight-Resonance Method'' is such a tool. It directly measures the anisotropies of the high-dimensional distribution of the vector activations, which are typically not visualisable. It is simple, robust and generalisable to any artificial neural network. It hopefully provides compelling evidence that artificial neural networks tend to organise their embeddings about these anisotropies and can be used to determine whether neurons respond to individual meanings. The tool may be seen as a generalisation and extension to previous works \citep{Szegedy2014, Bau2017}, borrowing the rotating basis of \citet{Bau2017}, but can be applied to any neural network where a privileged basis is suspected. It can capture a more holistic determination of the anisotropies, as it works across the full domain rather than just the positive activations, which previous methods have been limited by \citep{Bau2017}. Its application is flexible to individual or all privileged bivectors, which gives a local or global impression of the distribution. Moreover, parameters allow one to probe the angular distribution of embeddings at various angular scales for further insight. Therefore, this tool is hoped to be a singularly useful method in determining how models represent embedded data. 

    The results presented will establish whether the entire dataset produces this alignment since it naturally contains all subsets of human-interpretable concepts in the dataset. Therefore, global alignment would suggest local coding but not be definitive. To truly establish the presence of a local coding-like arrangement would require subdividing the dataset into categories reflecting meaningful human concepts and then observing whether each category has corresponding individual neurons (defined by decomposition in the privileged basis). Due to subjectivity, it is difficult to decide what constitutes a meaningful concept. Nevertheless, the proposed tool can be used in both circumstances.

    The principle of the work is simple, and the following analogy suitably describes it: \emph{it is as if one is counting the number of dust particles illuminated by a cone of light produced by a spotlight in a dark room. The spotlight completes full rotations, averaged across all privileged planes. If those particles have a tendency towards the corners of the walls. Then, when rotating, a resultant oscillation is observed in particle density with a frequency corresponding to the angular distribution of wall corners. Thus, it can be concluded that the shape of the room influences the dust distribution}. Hence, this method will be termed ``The Spotlight-Resonance Method'' (SRM). When transferring this analogy back to deep learning: the dust particles are each embedded activation vectors corresponding to a particular sample, whilst the corners of the walls correspond to privileged basis vectors and oscillations indicate that activations align with the privileged basis vectors - this tendency has been observed in the literature. If no such oscillations are observed by SRM, but instead a consistent signal, then it is unambiguously concluded that no activation distribution skewing occurs towards the privileged basis, and therefore, neurons probably do not correspond to concepts. This technique can also be performed across various subsets of the dataset, which may be expected to correspond to human-interpretable meanings, providing crucial evidence. In this paper, this methodology is discussed, along with some examples of SRM applied to small neural networks. It is hoped that the tool's simplicity, easy interpretability, and versatility can then find applications within the wider deep learning field, which will serve as a stepping stone to answering this fundamental question of representational alignment.
\section{Methodology}
   Below are two steps required to produce the method. \textit{Section}~\ref{Sec:Spotlight} explains everything required to implement the Spotlight-Resonance method for any artificial neural network, with detailed reasoning. \textit{Section}~\ref{Sec:ModelSpecifics} gives some essential considerations for models used.
   
   A quick-to-implement summary of \textit{Section}~\ref{Sec:Spotlight} is provided in \textit{App.}~\ref{App:BarebonesImplementation} without the accompanying mathematical justification.

   \subsection{The Spotlight-Resonance Method \label{Sec:Spotlight}}

   Implementing the SRM method is broken down into two further steps. First, a $n$-dimensional rotation matrix is generated, which rotates within a desired plane. Second, that rotation matrix is used to perform the Spotlight-Resonance method. There are many ways to generate such rotations, but the one discussed is reasonably simple to implement.

   \subsubsection{Generating Rotation Matrices using privileged Bivectors}

   The Spotlight-Resonance method is calculated across all privileged planes at any particular layer in the model. For this explanation, there are $n$-neurons in the particular layer, so a $\mathbb{R}^n$ activation space. Generalising, there may be $m$-privileged basis unit-vectors, denoted $\hat{b}_i$, induced by a functional form choice within this space --- note this basis can be under/overcomplete too. A privileged plane, to be termed a privileged bivector, is defined by the plane produced by two distinct privileged basis vectors: $\hat{b}_i\in\mathbb{R}^n$ and $\hat{b}_j\in\mathbb{R}^n$ measured from a third point: the origin, $\vec{0}$. The `spotlight' is then rotated in each privileged bivector plane one at a time for a complete rotation.

   In three dimensions, a cross-product could be utilised as an axis of rotation normal to a plane since there are coincidentally three basis bivectors and three basis vectors scaling as $m$ and $0.5m\left(m-1\right)$ respectively. Yet the cross-product is limited to three dimensions due to this coincidence. Instead, the wedge product allows this concept to be generalised to $m$ basis vectors and is therefore necessary for arbitrary privileged bases. A bivector is an oriented plane defined by the wedge (or exterior) product of two vectors. This method restricts the wedge product to two non-identical basis unit vectors. The bivectors are required to produce the matrix rotations needed for the method in the plane defined by the two chosen basis vectors.

   The privileged vectors form the set $\{\hat{b}_i|i\in\left[0,1,\cdots,m-1\right]\}$, whilst the privileged bivectors form the set $\mathbb{B}=\{\hat{b}_i\wedge \hat{b}_j|\left(i\neq j\right)\cap \left(i,j \in \left[0,1,\cdots,m-1\right]\right)\}$. The latter can be a set of unordered or ordered pairs termed \textit{Permutation-SRM} or \textit{Combination-SRM} respectively, depending on the user's symmetrisation preference for the later plot\footnote{There are several additional design choices so far: one can also produce privileged bivectors defined by three points, corresponding to three privileged basis vectors --- this can be more appropriate for simplexes with three-fold discrete rotational symmetry. Furthermore, one may choose the bivectors to be constructed from only neighbouring basis vectors. This restriction was not used in this method but may be desirable.}. Each of these potentially non-standard, privileged basis bivectors can then be decomposed into the \textit{standard} bivector basis for $\mathbb{R}^n$, such that they become antisymmetric $\mathbb{R}^{n\times n}$ matrices. In practice, this is achieved as $\mathbb{B}\ni \mathbf{B}_{\alpha \beta}=\frac{1}{2}(\hat{b}_{\alpha}\hat{b}_{\beta}^T-\hat{b}_{\beta}\hat{b}_{\alpha}^T)$ for two basis unit-vectors $\hat{b}_{\alpha}, \hat{b}_{\beta}\in\mathbb{R}^n$.
   
   This antisymmetric-matrix-represented bivector can then be treated as a member of the special orthogonal Lie algebra $\mathfrak{so}\left(n\right)$, which can be used to generate rotations through the exponential map. In effect, exponentiating this bivector matrix results in an $n$-dimensional rotation matrix for a rotation in the plane defined by that bivector, as desired. In practice, this can be easily achieved by eigendecomposition of the matrix bivector, $\mathbf{B}_{\alpha\beta}=\sum_{i=0}^{n-1}\vec{v}_i\lambda_i\vec{v}_i^\dagger$ as shown in \textit{Eqn.}~\ref{Eqn:ExponentialMap}, where dagger indicates the hermitian conjugate. The eigendecomposition produces two non-zero conjugate eigenvalues; these are normalised to $\pm i$ so that $\theta=2\pi$ is one complete rotation: $\mathbf{R}\left(0\right)=\mathbf{R}\left(2\pi\right)=\mathbf{I}_{n\times n}$.
   \begin{equation}
      \operatorname{SO}\left(n\right)\ni \mathbf{R}_{\alpha \beta}\left(\theta\right) = \sum_{i=0}^{n-1}\vec{v}_i\exp\left({\theta \lambda_i}\right)\vec{v}_i^\dagger
      \label{Eqn:ExponentialMap}
   \end{equation}

   \subsubsection{Using Rotation Matrices for Spotlight Method}

   This step intends to have a vector within the plane rotate with $\theta$ --- this vector acts as the direction of the `spotlight'. It is achieved by pre-multiplying the basis vector with its corresponding rotation matrix: $\hat{b}_{\alpha}'\left(\theta\right)=\mathbf{R}_{\alpha\beta}\left(\theta\right)\hat{b}_{\alpha}$. Then, taking the vector embeddings of a (sub)set of the training or testing dataset at the desired layer, $\forall\vec{d}\in\mathcal{D}_{L}\subset\mathbb{R}^n$, find all unit-normalised activation vectors, $\hat{d}$, which are within angle $\phi$ of the reference vector. This is equivalent to those vectors which meet the following dot-product condition: $\hat{d}\cdot\hat{b}\geq \epsilon$, where $\cos\phi = \epsilon$. The quantity of interest is the ratio of the cardinality of the set meeting this condition to the cardinality of the original set $\mathcal{D}_{L}$, this is expressed in \textit{Eqn.}~\ref{Eqn:Measure} below. 
   \begin{equation}
      f_{\text{SRM}}\left(\theta; \mathcal{D}_{L}, \epsilon, \left\{\alpha, \beta\right\}\right)
    =\frac{\left|\left\{ \vec{d}\in\mathcal{D}_{L} \,|\, \hat{d}^T\mathbf{R}_{\alpha\beta}\left(\theta\right)\hat{b}_{\alpha}\geq\epsilon\right\}\right|}{\left|\mathcal{D}_L\right|}
      \label{Eqn:Measure}
   \end{equation}
   Varying the angle of `the spotlight' can allow for finer resolution of angular scales; however, this also reduces the number of data points producing the signal. If desirable, this formulation can be adapted to non-Euclidean geometries through the inner product.
   
   The method is then performed for all values of $\alpha$ and $\beta$ in the privileged basis, with the results collated. The expectation value for this quantity can be found in \textit{App.}~\ref{App:nBall}. Collation of the results could be achieved using an ensemble line plot, median, mean, or alternative method, depending on what is being measured. In this paper's results, an ensemble line plot and mean line are presented. 
 
   \subsection{Model and Training \label{Sec:ModelSpecifics}}

   There may be many contributions within a network to the privileging of a particular basis, alongside just activation functions. Many parts of the model may privilege their own respective bases, which may result in interference between multiple privileged bases and yield an overall global privileged basis. Some of these model choices include initialisations, normalisations, regularisations, optimisers, activation functions, and even the desired output layer structure. The hierarchy of these contributions is presently unclear, and in future studies, this technique could establish how each function influences the privileging of a basis. This complex interference effect may result in the observed tendency towards a particular basis. This interference may explain the imperfect alignments sometimes observed \citep{Olah2019}, or perhaps the imperfect alignment is a consequence of the shape of the non-linearity for beneficial computation --- which could be tested using alternative activation functions. In this work only the activation function's role in basis privileging is demonstrated, this is to demonstrate that it is functional form's basis privileging which is a direct cause of observed representational alignment. However, in a more general setting, these other sources could result in non-static and difficult-to-determine privileged bases which representations may then align to.

   This interplay of privileged bases initially complicates the test and establishing the SRM technique. Therefore, so-called isotropic choices will be used to minimise interference in all tests and results. This is discussed further in \textit{App.}~\ref{App:Isotropic}. In practice, this means minibatch momentum gradient descent, Xavier-\textit{normal} initialiser \citep{Glorot2010}, no regularisation or normalisation, and an autoencoding reconstruction task on MNIST \citep{Lecun2010} or CIFAR \citep{Alex2009}. These are essential training requirements when establishing the basic SRM method.  It is the latent layer of the autoencoder models which shall be analysed. All further details, such as model architecture and training specifics, are discussed in \textit{App.}~\ref{App:Architectures}. Overall, this serves to isolate activation functions as the sole contributor to the unambiguous privileged basis, which can be termed the: `activation function privileged basis' (if a single activation function is used across the network). The results of \textit{Sec.}~\ref{Sec:Results} use a model without an activation function before the latent layer. This removes the bounding of the activation function as a trivial confounding cause of any anisotropic distribution observed. Therefore, any results are solely due to only the change in encoder parameters due to all other factors being controlled. 

   Finally, a novel functional class of activation functions is used in all network models. It allows the privileged basis induced by the activation function to be varied in rotation and completeness by changing the number of vectors constituting it. In addition, it ensures that the privileged bases seldom coincide with the standard basis. This shows SRM's versatility on differing bases and demonstrates that basis alignment is due to functional form choices' inducements of privileged bases and not fundamental to the standard basis. Further details of its implementation are provided in \textit{App.}~\ref{App:GeneralTanh}.

\section{Results}
\label{Sec:Results}

\begin{figure}[htb]
       \begin{center}
       \includegraphics[width=0.9\textwidth]{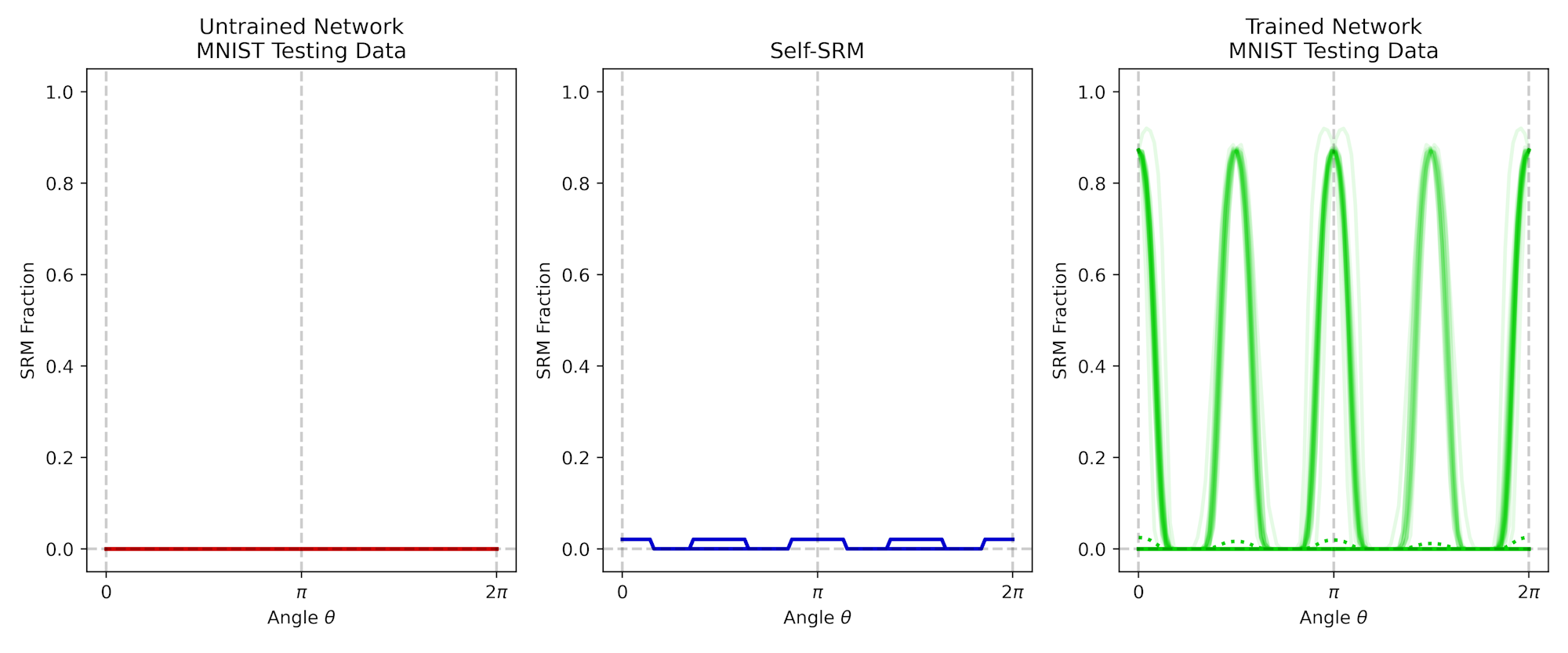}
       \end{center}
       \caption{Shows a representative example of combination-SRM applied to the small MNIST model with $n=24$ neurons and $m=48$ privileged basis vectors --- this creates a non-standard privileged basis along which elementwise tanh is applied. The spotlight angle used is $\epsilon=0.9$. An unseen testing data split was used to evaluate the SRM. The very faint, solid lines demonstrate the SRM fraction for each of the privileged bivectors (an ensemble line plot). Many of these translucent lines overlay, creating the dense oscillation pattern observed. The single dashed line per plot is the mean result across all privileged bivectors. The left plot shows the results of SRM for the network before training, whilst the right shows the exact same network after training. Centre shows self-SRM, which is SRM computed for the vectors of the privileged basis; this indicates what a local coding oscillation may appear like as a reference signal. It demonstrates that only after the training does the SRM oscillations become consistent with representations being axis-aligned with the privileged basis. Therefore, it appears training results in an activation symmetry breaking induced by the symmetry breaking functional forms. Due to page restrictions, further results can be found in \textit{App.}~\ref{App:ExtraTests}.}
       \label{Fig:PrimaryResult}
    \end{figure}

    A wide range of networks were tested with \textit{Fig.}~\ref{Fig:PrimaryResult} being a good representative example, further results can be found in \textit{App.}~\ref{App:ExtraTests}. Figure~\ref{Fig:PrimaryResult} demonstrates the combined-SRM method on the small MNIST autoencoder model. If the activation function induces a privileged basis-aligned representation after training, then a strong oscillation will be observed in the ensemble. This oscillation would be in phase with the reference self-SRM oscillation, indicating alignment.
    
    Observed in \textit{Fig.}~\ref{Fig:PrimaryResult} are clear oscillations in the SRM measure, only after training, which are in phase with the self-SRM. This can only be caused by an increased density of embedded representations in angles close to those privileged by the basis. The mean results, shown in dashed lines, for the trained network SRM and self-SRM closely match agreeing with this assertion. Meanwhile, the SRM values for the untrained network are very low. This is because there is a much greater proportion of space outside the spotlight cone than inside, so if representations are approximately uniformly distributed by random initialisation, then there is a low incidence with the spotlight cone.

    Furthermore, there is a significant variation in oscillation amplitude in the trained network: some planes' SRM have large amplitudes of around $90 \%$ of the dataset, whilst many are close to zero. This could be interpreted as the embedded activations only clustering about a subset of privileged basis vectors whilst having a constant offset for others. The reasons for this are unclear but may be due to the bias' role in superposition interference or an excessive number of neurons --- requiring further investigation. Supplementary examples of SRM are demonstrated in \textit{App.}~\ref{App:ExtraTests}, alongside SRM on human-interpretable subsets of the dataset which find \textit{grandmother neurons}. Overall, the general trend across all results is that representations align (or anti-align) with the privileged basis, solely caused by the activation function applied. No alignment is observed with the standard basis when a non-standard basis is privileged. 
\section{Conclusion}

    In conclusion, in all models tested, embedded representations tend to align (or anti-align) about the privileged basis vectors. These activations cluster around these significant directions and produce a change in distribution density which is shown to be directly detectable using the new SRM technique. These privileged directions are the extrema of the anisotropies caused by the applied functional forms. In these experiments, this can only be produced by the \textit{choice} of the activation function, demonstrating a clear cause and effect for the observed representational alignment. This provides strong supporting evidence for the observed tendencies of activations to cluster about the standard basis in prior works \citep{Vondrick2016, Bau2017, Olah2019, Elhage2022} --- since elementwise functional forms are used, which privilege the standard basis. However, this paper's results also establish that this clustering is actually around the privileged basis which is not necessarily the standard basis, as often thought. This is because the observed oscillations align with the privileged basis, but not with the standard basis when a functional form with non-standard basis privileging is implemented. Hence, prior observations of representations aligning with (standard) neurons are not an innate phenomenon of deep learning, but specifically due to choices in activation function functional forms. This demonstrates there is little significance behind the standard basis besides the current practice of using it to apply activation functions elementwise along. Instead, the more general concept of a privileged basis is shown to be the fundamental quantity.
    
    This sets a foundation for a new generation of neural network functional forms, which may extend beyond activation functions. These can be used to directly influence the representational alignment in desirable and measurable ways for different tasks. Several (non-standard) grandmother neurons are also identified in \textit{App.}~\ref{App:ExtraTests}, which seem to respond to human-interpretable concepts anywhere in the provided image. This is surprising since the architectures are fully connected feedforward networks opposed to the translational equivariant convolutional networks, where this behaviour may be expected. The non-standard alignment also supports the hypothesis of general linear features. Although results are demonstrated on autoencoders for establishing the technique, the methodology is general and can be applied to all known deep learning models. Therefore, SRM may also be used to add evidence on the neural collapse phenomena \citep{Papyan2020}, which may be a privileging of a basis by the choice of one-hot output. Future work could expand this analysis on the functional form hierarchy of basis privileging as well as more thoroughly investigating local coding through meaningful subsets of the dataset. 
    
    Moreover, this paper establishes the spotlight-resonance method, in its various forms, as a simple, interpretable, versatile and powerful tool for establishing representational alignment in general deep learning models.

\newpage
\bibliography{references.bib}
\bibliographystyle{iclr2025_conference}
\newpage
\appendix

\section{Summary for Implementing SRM}
\label{App:BarebonesImplementation}

   All the steps for SRM can be summarised as follows for easier implementation.

   Initially, rotation matrices for each (privileged) bivector must be generated:
   \begin{enumerate}
      \item Determine the privileged basis vectors $\hat{b}_{i}$ corresponding to the latent layer to be analysed.
      \item Calculate all pairwise privileged basis bivectors in matrix form $\mathbf{B}_{\alpha \beta}=\frac{1}{2}\left(\hat{b}_{\alpha}\hat{b}_{\beta}^T-\hat{b}_{\beta}\hat{b}_{\alpha}^T\right)$ and $\alpha\neq\beta$. Decide in this step whether to use permutation-SRM, combination-SRM or other variations.
      \item Eigendecompose each bivector $\mathbf{B}_{\alpha \beta}=\sum_{i=0}^{n-1}\vec{v}_i\lambda_i\vec{v}_i^\dagger$.
      \item Normalise the two non-zero conjugate eigenvalues to $\pm i$.
      \item Generate in-plane rotation by exponentiation of those eigenvalues with angle $\theta$, as shown in \textit{Eqn.}~\ref{Eqn:ExponentialMapAgain}.
   \end{enumerate}
    \begin{equation}
      \operatorname{SO}\left(n\right)\ni \mathbf{R}_{\alpha \beta}\left(\theta\right) = \sum_{i=0}^{n-1}\vec{v}_i\exp\left({\theta \lambda_i}\right)\vec{v}_i^\dagger
      \label{Eqn:ExponentialMapAgain}
   \end{equation}
    For finding vectors within the spotlight:
    \begin{enumerate}
        \item Forward pass a (sub)set of $d$-samples of the dataset to the $n$-neuron latent layer, which is to be analysed. Each sample can be stacked into the matrix: $\mathbf{A}\in\mathbb{R}^{d\times n}$.
        \item Normalise this matrix row-wise. This requires calculating the 2-norm in $\mathbb{R}^n$ of each row in the above matrix. This ensures all the stacked vectors now become stacked unit vectors.
        \item Rotate the corresponding privileged basis unit vector with its plane rotation: $\mathbb{R}^n\ni\hat{b}_{\alpha}'\left(\theta\right)=\mathbf{R}_{\alpha\beta}\left(\theta\right)\hat{b}_{\alpha}$.
        \item Take the dot-products between rows of the matrix $\mathbf{A}$ and the plane-rotated privileged vector $\hat{b}_{\alpha}'$. This produces a vector of similarities $\left[-1, 1\right]^d$.
        \item Count the number of elements of the vector which are greater than the threshold $\epsilon$.
        \item Divide this number by $d$ to produce the final SRM value for the current $\alpha$, $\beta$ and $\theta$.
        \item Repeat steps two through five for all $\alpha$'s, $\beta$'s and $\theta$'s to be tested. 
        \item Plot as means, medians or each sample of $\alpha$ and $\beta$ (ensemble plot), across the dependent variable $\theta$.
    \end{enumerate}
    This concludes the basic implementation. The full code implementation is available at the GitHub link \url{https://github.com/GeorgeBird1/Spotlight-Resonance-Method}.
\newpage  
\section{Extra Tests\label{App:ExtraTests}}
    In the following subsections, additional results for the Spotlight Resonance method are provided.

    The first section (\textit{Sec.}~\ref{Sec:PrivUnique}) demonstrates that the observed alignment phenomena are unique to the privileged basis. It compares otherwise identical SRM tests using a random basis, the standard basis and the privileged basis. It is found that only the privileged basis produces a signal. This additionally justifies that analysis along the privileged is most salient when determining alignment.

    The second section (\textit{Sec.}~\ref{Sec:LocalCoding}) provides evidence that grandmother neurons are present in several networks tested. These are found to respond to sea/sky, vehicles, and eyes in the large CIFAR network tested. This is additionally interesting since these are not convolutional networks being tested, so they don't feature translational equivariance. Yet, they still seem to detect/represent localised objects, such as eyes, present across the image. Results on an MNIST autoencoder are also provided. This is preliminary evidence that representations aligned with certain (privileged) neurons represent human-interpretable concepts.

    The third section (\textit{Sec.}~\ref{Sec:Elementwise}) provides further supporting results for conclusions reached in \textit{Sec.}~\ref{Sec:Results}. These are for an elementwise basis, with results continuing to show that representations tend to align with basis directions. Therefore, this shows the repeatability of the observation across various networks and model architectures. In this section, SRM is performed on larger networks, which produce differing strength oscillations depending on the plane.

    In \textit{Sec.}~\ref{Sec:Simplex} and \textit{Sec.}~\ref{Sec:Overcomplete}, results are shown for simplex and overcomplete activation function bases, respectively. This demonstrates the versatility of the SRM technique. It also presents highly unusual cases where activation functions are not applied typically. These results can offer further insights into the fundamental behaviour of deep learning models. For example, the representations for the simplex basis are consistently anti-aligned, whilst the overcomplete basis varies between alignment and anti-alignment. To the best of the author's knowledge, this is also the first time the effects of varying the (privileged) basis of activation functions have been studied.

\subsection{Is it Unique to Privileged Bases?}
\label{Sec:PrivUnique}

    The prior results state that SRM produces a signal after a model's training that is consistent with axis-aligned representations with the privileged basis but not necessarily the standard basis. This section will evidence this statement. 
    
    The findings also support performing SRM only on the privileged basis-bivectors since this is sufficient for determining axis alignment. Therefore, only performing SRM on the privileged basis gives a good holistic impression of the overall angular distribution. It further demonstrates that SRM only produces a positive oscillating signal when there is clear alignment or anti-alignment present, with no signal otherwise. In future work, this may enable early stopping techniques to determine when training is complete.

\begin{figure}[H]
       \begin{center}
       \includegraphics[width=0.9\textwidth]{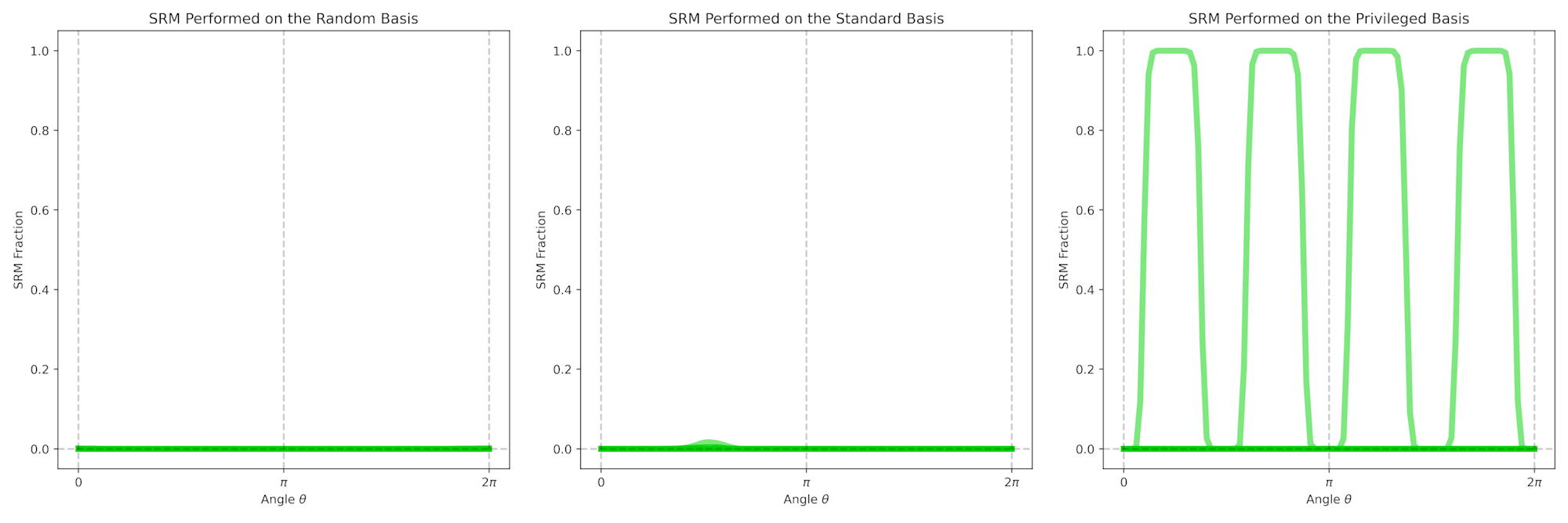}
       \end{center}
       \caption{The left plot demonstrates combination-SRM performed using random normal basis-bivectors, the centre shows combination-SRM using bivectors of the standard basis, whilst the right plot shows combination-SRM for the bivectors of the privileged basis. Every other parameter was kept constant across all plots, and in all cases, the number of basis vectors was equal. The network tested was the trained small MNIST model $n=10$, $m=20$ evaluated on the MNIST test set split. These results were consistent with those of other networks tested. A value of $\epsilon=0.9$ was used. The small peak on the centre plot could be for several reasons, such as a coincidently close alignment between a standard bivector and a privileged bivector or a very small subset of representations which do not display privileged basis axis alignment. In either case, the signal is very small, not oscillating or in phase with the standard basis, so can be considered insignificant.}
       \label{Fig:VariousBases}
    \end{figure}

    Figure~\ref{Fig:VariousBases} shows SRM performed on three differing bases: a random basis, the standard basis and the activation function privileged basis.

    This demonstrates that the axis alignment is unique to the privileged basis and that performing SRM on only the privileged basis is sufficient for capturing the axis alignment of representations in the model. 

    Since the behaviour is unique to the privileged basis, it can firmly be established that there is nothing innately special about the standard basis. Any representational alignment is, therefore, due to basis-dependent functional forms, which are anisotropic and consequently induce anisotropy in the embedded activations. In this case, the only privileging functional form was the activation function, allowing a definitive privileged basis to be established.
\subsection{Local Coding Results}
\label{Sec:LocalCoding}
    The primary motivation for developing the spotlight resonance method was to determine whether local coding (aligned with a privileged basis) is present in a general network. The workshop paper is intended to showcase the SRM method; however, in this section, preliminary results for local coding are discussed. These preliminary results are provided for the CIFAR and MNIST datasets, showing that directions corresponding to the privileged basis have a variation in activation embedding, which meaningfully represents human-interpretable concepts. These are for autoencoders trained in reconstruction, so no specific labelling is induced by a classification layer. It is shown that individual neurons corresponding to the \textit{privileged} basis do represent meaningful subdivisions in the datasets even for purely self-supervised tasks.
    
    All other results have been evaluated across the whole testing set, clearly demonstrating alignment in general with the privileged basis. However, to determine the presence of locally coded neurons or grandmother neurons, the individual neurons must correspond to human-identifiable classifications. This is essential, as even if the whole dataset's SRM is uniform, subsets could still oscillate about certain privileged basis planes, indicating local coding-like behaviour. Consequently, varying alignment must be demonstrated for various meaningfully partitioned subsets of the whole dataset. Ideally, a dataset such as Broden should be used, as produced by \citet{Bau2017}, but this additional analysis was out-of-scope for this paper. Instead, subsets corresponding to individual digits of the MNIST dataset are shown and various classifications within CIFAR. Furthermore, oscillations could also indicate differing codings, such as higher frequency oscillations for sparse coding. An alternative self-SRM could be constructed for such a test, but it was out of scope for this workshop paper.

    With a sufficiently trained CIFAR network, regions of the embedded activations may be expected to represent subdivisions of the human-labelled categories, such as colours of trucks, types of dogs, etc. On a less granular scale, a network may reproduce the human-labelled classifications: cars, trucks, aeroplanes, etc. On a yet larger scale, it may be expected that representations are organised into broader concepts such as the sky, water, blue, and the presence of roads. When analysing the large CIFAR network, specific neurons are found for these broad categories, which also have a variation of response with the human labelling. Therefore, it can be concluded that there are individual neurons (of the privileged basis) which do represent single ideas, so effectively grandmother neurons. However, these are analyses of individual neurons, so an overall local coding cannot be established from these results, especially since some neurons did not show such a variation in response per category when analysed. This suggests grandmother neurons are present in the network but not universal.

\begin{figure}[htb]
       \begin{center}
       \includegraphics[width=0.99\textwidth]{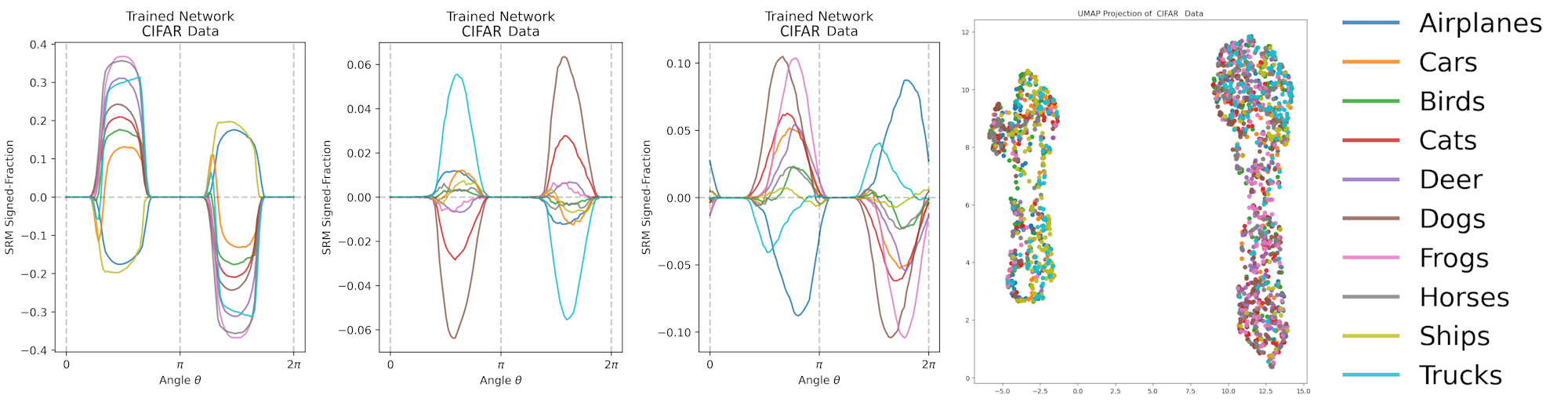}
       \end{center}
       \caption{The three leftmost plots show the signed spotlight resonance method performed on handpicked single privileged bivectors. All three indicate that representations are strongly aligned with a single neuron, and the sign and strength of firing of that neuron represent a human-interpretable meaning, as discussed below. Therefore, all three are likely \textit{grandmother neurons} for the CIFAR dataset. The second-to-rightmost plot is a UMAP \citep{McInnes2020} embedding of the latent layer, whilst the rightmost is a colour-coded key for the diagram. None of these observed oscillations in SRM were present before training. For the experiment, $\epsilon=0.75$ was used.}
       \label{Fig:CIFARlc}
    \end{figure}

    The Signed Spotlight Resonance method is described by \textit{Eqn}~\ref{Eqn:SignedSpotlight}. Intuitively, it is like the standard SRM but subtracts off activations within the negative direction cone from the positive direction cone.
    \begin{align}
      f_{\text{signed-srm}} &\left(\theta; \mathcal{D}_{L}, \epsilon, \left\{\alpha, \beta\right\}\right) \nonumber
      \\[1.5mm]
      &=\frac{\left|\left\{ \vec{d}\in\mathcal{D}_{L} \,|\, \hat{d}^T\mathbf{R}_{\alpha\beta}\left(\theta\right)\hat{b}_{\alpha}\geq\epsilon\right\}\right|-\left|\left\{ \vec{d}\in\mathcal{D}_{L} \,|\, \hat{d}^T\mathbf{R}_{\alpha\beta}\left(\theta\right)\hat{b}_{\alpha}\leq-\epsilon\right\}\right|}{\left|\mathcal{D}_L\right|}
      \label{Eqn:SignedSpotlight}
   \end{align}
    To begin, \textit{Fig.}~\ref{Fig:CIFARlc} shows local coding on a subset of latent layer neurons in the large CIFAR network.
    The leftmost plot shows two oscillations at $\pi/2$ and $3\pi/2$, therefore corresponding to the same neuron (decomposed in privileged basis) in opposing directions. When observing how it varies across classes, it is strongly negative firing for the ships, aeroplanes and slightly for the car categories (in order of peak magnitude), whilst positive firing for frogs, horses, trucks, deer, dogs, cats, birds and cars (in order of peak magnitude). Subjectively, this seems to represent the presence of woodland scenes and the absence of sky and water. Looking at samples of the dataset, this is rather intuitive as ships are rarely pictured out of the water, aeroplanes are mostly pictured in the sky or with substantial sky in them \textit{but} infrequently pictured on a runway. Frogs are nearly always pictured in green, swamp-like backgrounds, horses and deer in fields, dogs and cats sometimes in the wild but often in human environments, with birds appearing on the ground and with sky backgrounds and similar for cars. Therefore, this leftmost neuron appears to be distinctly a scene detector and separates the human-labelled categories into proportions, reflecting how much of the sky or water is typically viewable in samples of that categorisation. It also appears this neuron approximately represents the horizontal separation observed in the UMAP plot. Additional neurons very similar to this one in response were found numerous times in the network, consistently strongly axis-aligned like this one. This infers a redundancy to this specific detector. The consistent axis alignment suggests that it is not just an oscillation in a linear direction which happens to also cause oscillations along the privileged bases when projected but instead suggests several neurons individually responding to similar stimuli. The reasons for this are presently unclear, especially the cause for this observed redundancy. Yet, it may also be indicative of sparse coding due to the multiple redundancy, but this is not conclusive.  

    The second-from-leftmost plot shows representations strongly aligned with a single neuron at $\pi\pm\pi/2$, responding strongly negative to dogs, then cats, then slightly to frogs and deers. It responds most positively to trucks, then cars, then aeroplanes, then ships and less so to horses and birds. One may subjectively interpret this as a detector for mechanical vehicles, metal or grey, whilst negative firing for human homes with pets. Often, the pets are taken with professional photography backgrounds or at home, unlike horses, frogs, deer and birds, to which the neuron responds little. It is difficult to tell whether this has a corresponding direction in the associated UMAP latent space embedding plot.

    Finally, the third-to-leftmost plot has representations generally aligned with the single neuron at $\pi\pm\pi/2$. This neuron strongly negatively activates for aeroplanes, then trucks, with positive activations in the greatest magnitude order of dogs, then frogs, cats, deer, cars, birds, and horses --- finally, with a little activation for ships. This seems like a detector similar to the leftmost plot but with some differences. The presence of animals and cars in the positive activation could suggest that it is a round-eye-like object to which it responds. Dogs, cats and frogs seem to often be imaged from the front, with eyes visible, whilst deers, birds and horses are less so (and often from further away, so smaller apparent eyes from the camera perspective). The headlights of a car could be mistaken by the network for eyes. Therefore, subjectively, this neuron might be effectively the presence-of-eyes neuron. This neuron does not seem to clearly correspond to a direction in the UMAP plot.

    Therefore, it can be preliminarily concluded that \textit{some} neurons in the latent layer of a large CIFAR network do respond uniquely to human-interpretable categories, with embedded activations strongly aligning with these specific neurons. This suggests that `grandmother neurons' are spontaneously produced in a reconstruction task, but there is so far insufficient evidence to conclude a local coding across the full network. It is especially interesting since these are fully connected feed-forward networks, not convolutional. Therefore, the networks do not have translational equivariance but still seem to respond to the general presence of the stimulus in the image.

\begin{figure}[htb]
       \begin{center}
       \includegraphics[width=0.99\textwidth]{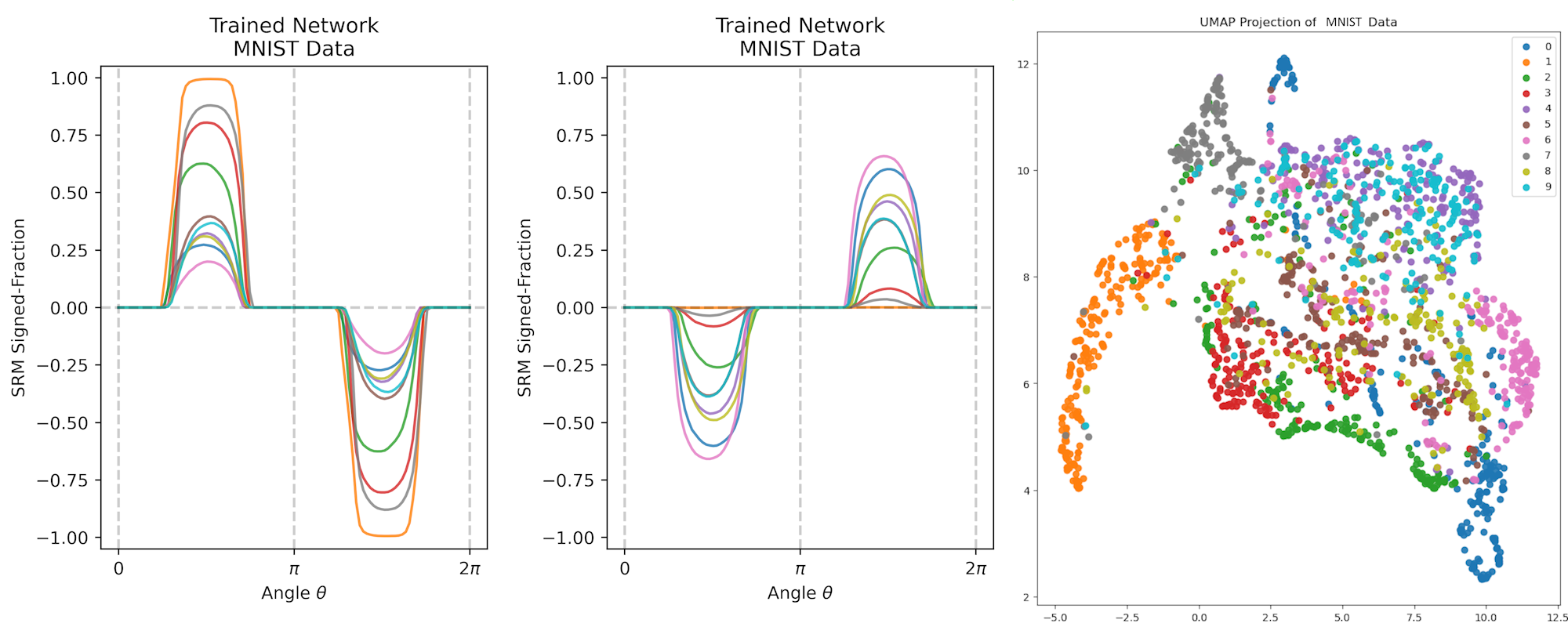}
       \end{center}
       \caption{The two leftmost plots show the signed spotlight resonance method performed on handpicked single privileged bivectors. Both indicate that representations are strongly aligned with a single neuron, and the sign and strength of firing of that neuron represents a human-interpretable meaning, as discussed below. Therefore, both are probably \textit{grandmother neurons}. The rightmost plot is a UMAP of the embedding of MNIST in the latent layer, and it includes a colour-coded key for the diagram. None of these observed oscillations in SRM were present before training. For the experiment, $\epsilon=0.75$ was used.}
       \label{Fig:MNISTlc}
    \end{figure}

    Similar results are seen in \textit{Fig.}~\ref{Fig:MNISTlc} for the large MNIST network.
    For the leftmost plot, there is strong representational alignment for a single (privileged basis) neuron activating positively for the digits 1, then 7, 3, and 2, then a significant gap followed by 5, 9, 4, 8, 0 and finally 6 --- ordered in most positive firing. This neuron is challenging to categorise its meaning, though roughly it appears to activate strongly for the presence of an upper leftward facing sharp $>$ or curved $\supset$ open shape to them. The central plot also has a strong representational alignment with a single privileged neuron, but this time has a strong positive activation for digits 6, 0, 8, 4, 9, 5, and 2, then a gap followed by 3 and hardly any activation for 7 or 1. This is the opposite ordering of the leftmost neuron, suggesting it responds to the absence of a leftward hook shape.
    
    Other random bases were chosen, and the signed-SRM produced no signal, concurring with \textit{Sec.}~\ref{Sec:PrivUnique}. Therefore, at least from these preliminary results, it can be concluded that deep learning models, to some degree, have some locally coded, or grandmother, neurons which represent distinct,  meaningful concepts to humans. A more thorough analysis using the spotlight resonance method should be undertaken to provide definitive evidence, particularly on datasets such as Broden \citep{Bau2017} with compelling human-interpretable subdivisions of the dataset.
    
\subsection{Further Elementwise Bases} \label{Sec:Elementwise}

\begin{figure}[htb]
       \begin{center}
       \includegraphics[width=0.9\textwidth]{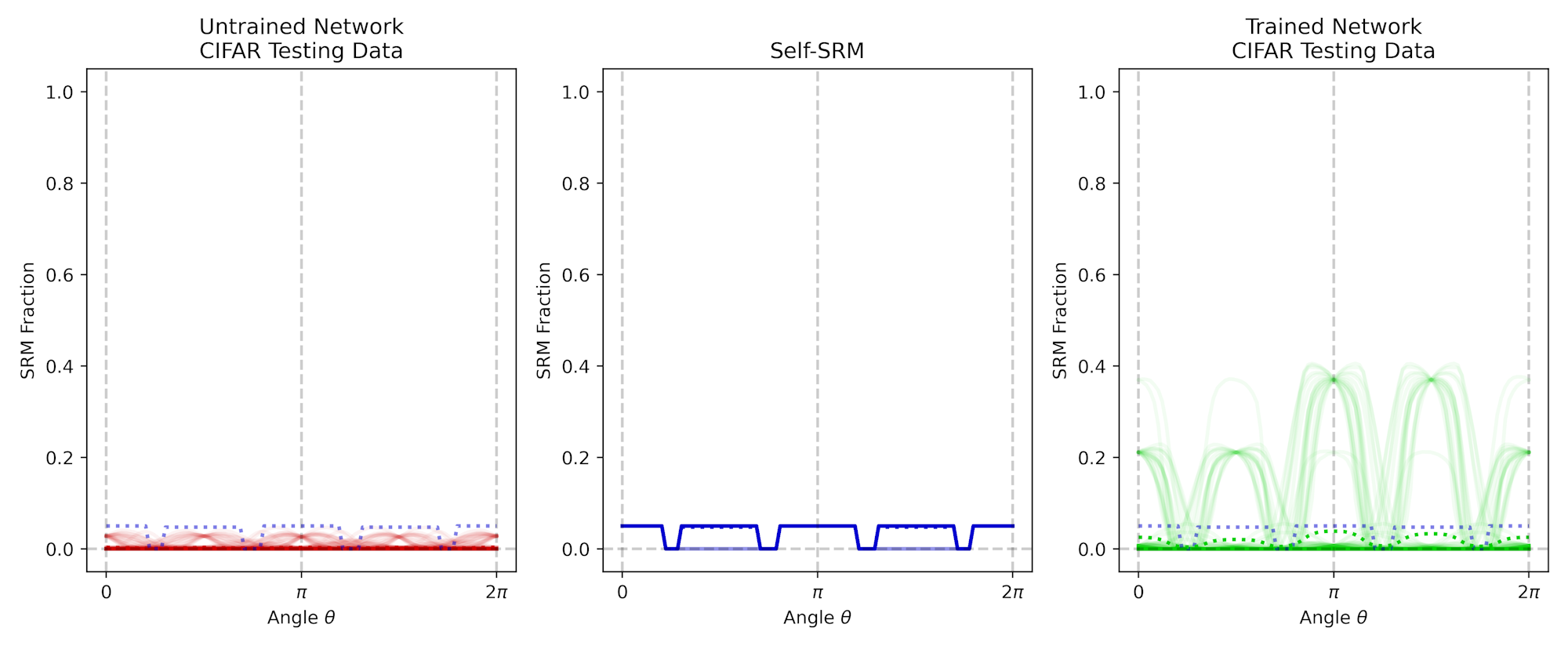}
       \end{center}
       \caption{The left plot shows SRM performed on the untrained large CIFAR network, whilst the right plot shows SRM on the same network following training on CIFAR, and the centre plot shows the self-SRM measure. A value of $\epsilon=0.8$ was used for combination SRM --- smaller than usual. The number of basis vectors is $m=20$ in $\mathbb{R}^{10}$, therefore elementwise. Although a smaller value of $\epsilon$ was necessitated to observe any signal, a strong basis alignment can be observed. This lower value of $\epsilon$ required suggests a more diffuse alignment with the privileged basis. In the untrained plot, there are several dense crossover points in the SRM value, which can be seen at $n\pi/2$. However, these are all small valued oscillations. This is not thought to be a signal, but due to the geometry of combination-SRM, discussed further below.}
       \label{Fig:ElementwiseOne}
    \end{figure}

    The vast majority of elementwise bases ($m=2n$) tested continued to show the basis-aligned signal found in \textit{Sec.}~\ref{Sec:Results}, with only a few networks having no SRM signal until smaller values of $\epsilon\approx0.7$ were chosen. These smaller value $\epsilon$ exceptions suggest that the activations are more diffuse but \textit{still} aligned with the privileged basis since the signal is only detectable with large spotlight-cone angles. This is suspected to be due to incomplete training of the networks. In all cases, the privileged basis does not coincide with the standard basis, so alignment is directly due to the functional form of the activation functions. \textit{Fig.}~\ref{Fig:ElementwiseOne} shows a `large CIFAR' network's SRM oscillation at a lowered value of $\epsilon=0.8$ --- the observed oscillations continue to support the conclusions reached but indicate a more diffuse alignment.

\begin{figure}[H]
       \begin{center}
       \includegraphics[width=0.9\textwidth]{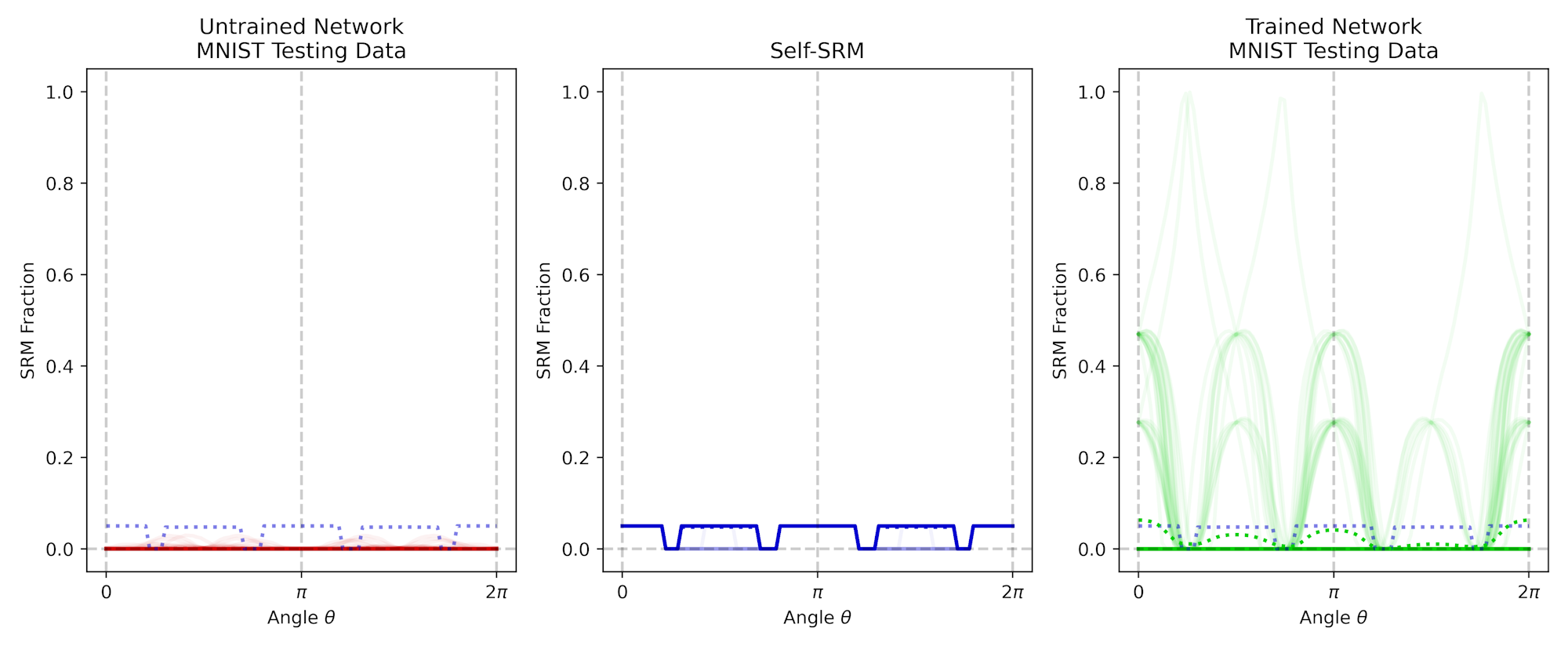}
       \end{center}
       \caption{The left plot shows SRM performed on the untrained large MNIST network, whilst the right plot shows SRM on the same network following training on MNIST, and the centre plot shows the self-SRM measure. A value of $\epsilon=0.8$ was used for combination SRM. The number of basis vectors is $m=20$ in $\mathbb{R}^{10}$. The oscillation in the trained data continues to strongly align with the privileged basis. A split can be observed in the SRM values for each peak, a lower and a higher amplitude alignment. This was found to be very common in the large autoencoder models (which include an activation function before the latent layer). On the left untrained model plot, the SRM values are much lower and generally uncorrelated with self-SRM. In the right plot, several anti-aligned oscillations are also observed.}
       \label{Fig:ElementwiseTwo}
    \end{figure}

    These larger autoencoder models continue to demonstrate alignment but often tend to show a separation in the SRM values. This is demonstrated in \textit{Fig.}~\ref{Fig:ElementwiseTwo}. This complicated structure likely emerges to benefit performance on the reconstruction. It may be indicative of local coding since it suggests differing subsets of data are being distributed unevenly across various privileged basis vectors. This more discerning embedding may be unique to larger networks, which can achieve better separation of contrasting features in the data.

    The low-valued dense crossover points, in the untrained plots of \textit{Figs.}~\ref{Fig:ElementwiseOne} and \ref{Fig:ElementwiseTwo}, probably should not be confused with a unique positive signal. This is because for all $\alpha$ values for bivector $\hat{B}_{\alpha \beta}$, the SRM values must agree at $\pi\pm\pi/2$. This is because these rotations always correspond to a spotlight pointing in direction $\mp\hat{b}_\beta$, whilst $\pm\hat{b}_\alpha$ for $n\pi$. This produces a denser region where ensemble values must cross regardless of a signal. This is corroborated by the mean SRM value not correlating with the self-SRM, unlike the trained plot. Thus, the crossover points are likely only an artefact of the geometry. Individual waves can be observed to be generally uncorrelated with the self-SRM before training but in phase with self-SRM after. However, a small number of waves are in phase, which may be due to the larger network having a bounded activation function before the latent layer. The bounding may result in a slight in-phase distribution, as large magnitude activations are reduced to form a hyper-cubic shape around the basis directions. Though the random initialised weights, after the activation function, may be expected to rotate this anisotropic distribution contrary to what is observed. Results from random matrix theory for the initialisation may explain this. Nevertheless, this seems to explain why the phenomenon does not occur in the smaller autoencoder models, though does need greater exploration in future work. 

    There are a few high-magnitude anti-aligned oscillations in the trained plot of \textit{Fig.}~\ref{Fig:ElementwiseTwo}. This is particularly interesting and is not typically observable in alternative methods to SRM. Several factors could result in this anti-aligned oscillation: perhaps it is an artefact of incomplete training, where activations are midway through crossing between two privileged basis vectors. Despite this, it would be unlikely to observe this crossover at the precise moment that it is perfectly anti-aligned; instead, it is probably beneficial to performance somehow. Perhaps it is representation capacity: if an elementwise basis is considered, with $2n$ privileged vectors, then representation alignment with the privileged basis (local coding for elementwise basis privileging) limits the representation capacity to the number of privileged vectors, $2n$. However, if representational anti-alignment is used (effectively a dense coding for elementwise basis privileging), then the representational capacity is $2^n$. However, this higher representation capacity comes at the cost of increased interference and challenges with disentangling the representations. Therefore, the observation may be consistent with sparse coding, where some representations are aligned and some anti-aligned, balancing these factors.
    
    Furthermore, this argument would suggest that smaller privileged bases $m\ll 2n$, might prefer anti-alignment, as this keeps higher representation capacity, compared to aligned representations, whilst also featuring less interference and disentangling challenges for the network due to the smaller number of privileged basis vectors being more angularly separated. This is consistent with the simplex basis results below. However, this argument requires further study and does little to explain the highly overcomplete basis observations. Overall, this shows that SRM can give a more nuanced insight into the alignment of data embeddings.

\subsection{Simplex Bases}
\label{Sec:Simplex}
    
\begin{figure}[htb]
       \begin{center}
       \includegraphics[width=0.9\textwidth]{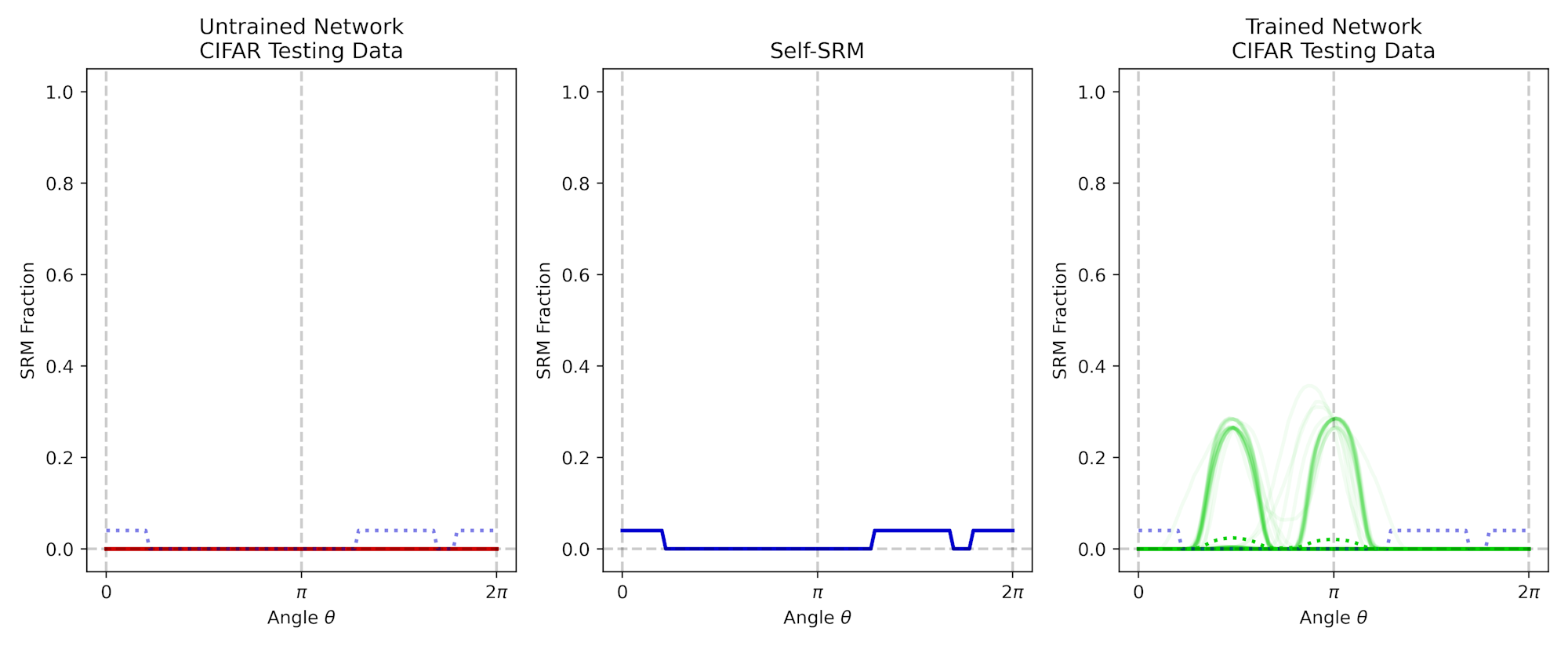}
       \end{center}
       \caption{The left plot shows SRM performed on the untrained small CIFAR network, whilst the right plot shows SRM on the same network following training on CIFAR, and the centre plot shows the self-SRM measure. A value of $\epsilon=0.8$ was used for combination SRM. The number of basis vectors is $m=25$ in $\mathbb{R}^{24}$. A clear oscillation is observed in the rightmost plot, which has a large amplitude at angles where self-SRM has a small amplitude and vice-versa. This strongly indicates that simplex bases cause an anisotropic distribution in the embedded activations, but in this specific case it is anti-aligned with the privileged basis.}
       \label{Fig:SimplexOne}
    \end{figure}

    The simplex bases are characterised by $m=n+1$ vectors uniformly angularly distributed in $\mathbb{R}^n$. When performing the SRM technique on such an activation function's privileged basis, an anti-basis aligned oscillation is observed. This is displayed in \textit{Fig.}~\ref{Fig:SimplexOne}.

    This anti-alignment was observed in all of the simplex bases tested for small MNIST and CIFAR autoencoder models but not the large variety (where alignment was observed). It demonstrates how the broken rotational symmetry, caused by the activation functional form, may induce a privileged basis at either extrema: maximally aligned or maximally anti-aligned - which may be highly dependent on the particular anisotropic non-linearity of the function. The reason for the contrasting alignment in the larger networks is unclear. If one wishes to manipulate the representation distribution in a particular way, it indicates that the choice of activation function could play a crucial role.
    
\subsection{Highly Overcomplete Bases}
\label{Sec:Overcomplete}

\begin{figure}[htb]
       \begin{center}
       \includegraphics[width=0.9\textwidth]{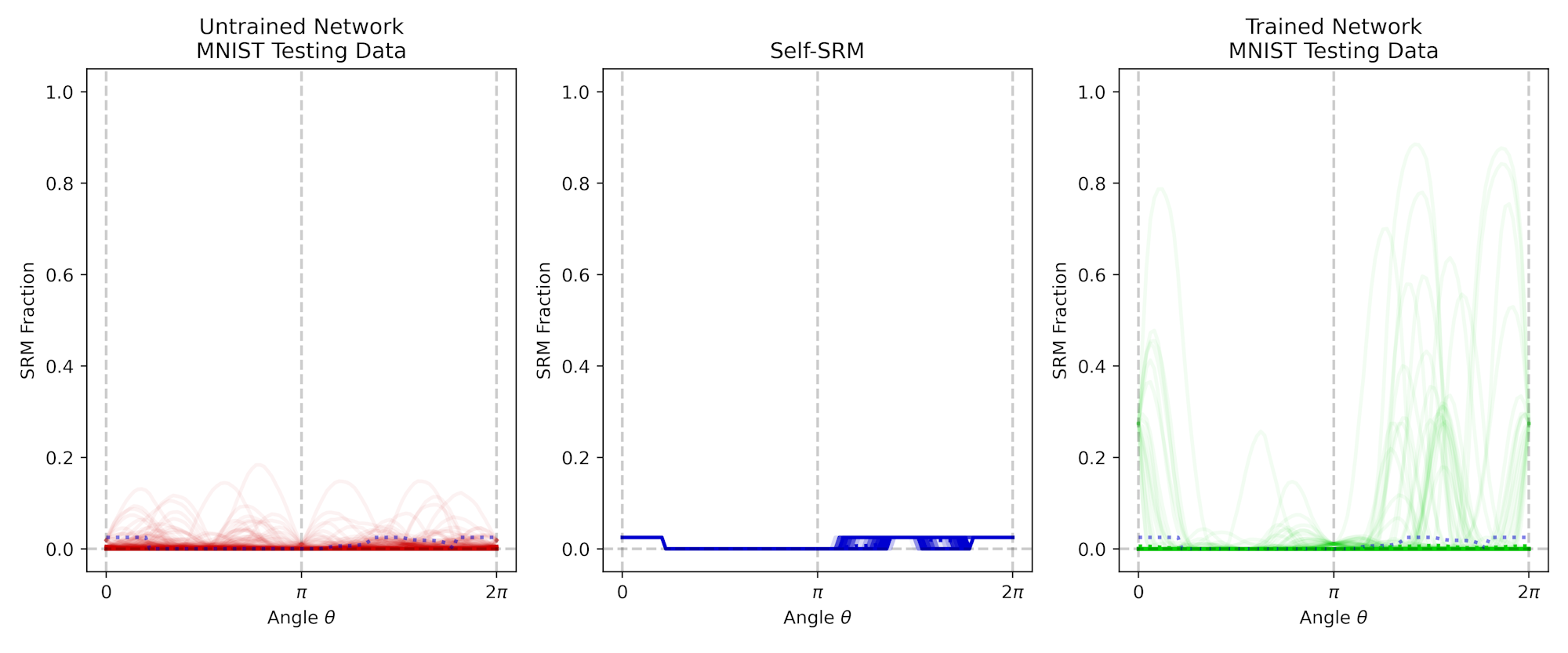}
       \end{center}
       \caption{Shows combination-SRM performed on a large MNIST model with $n=10$, $m=40$ and $\epsilon=0.8$. The left plot shows SRM performed on the untrained model, whilst the right plot shows the method performed on the trained model. The centre plot shows self-SRM for the $m=40$ basis vectors embedded in $\mathbb{R}^{10}$. In the trained plot, it can be observed that representations tend to align with a privileged basis, as the SRM on the trained model is similar to the self-SRM reference.}
       \label{Fig:Overcomplete1}
    \end{figure}

\begin{figure}[htb]
       \begin{center}
       \includegraphics[width=0.9\textwidth]{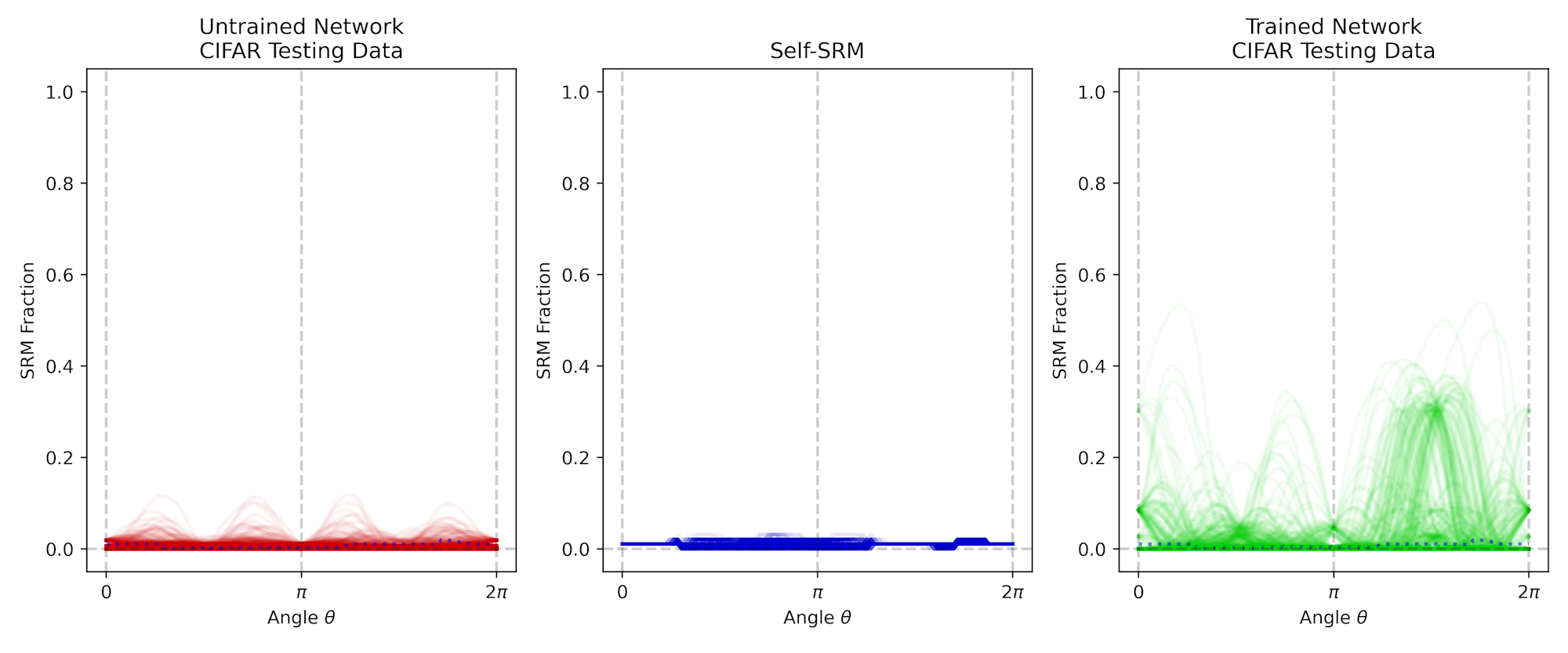}
       \end{center}
       \caption{Shows combination-SRM performed on a large MNIST model with $n=24$, $m=96$ and $\epsilon=0.6$. As before, the left plot shows combination-SRM performed on the untrained network, whilst the right plot shows it performed on the same network after training. The centre plot gives the self-SRM reference values to compare against the left and right plots. Notably, $\epsilon$ was required to be significantly lower before any signal was observed. It can be, therefore, concluded that the angular distribution of embeddings is very diffuse, in opposing directions to the privileged basis vectors. The lowered $\epsilon$ is likely the reason that a non-zero SRM is observed in the left untrained plot. This plot has four regular peaks forming an oscillation. The reason for this could be the bounded activation functions or possibly a geometrical artefact in the test.}
       \label{Fig:Overcomplete2}
    \end{figure}

    These results vary significantly across different networks, but all tend to have representations aligned or anti-aligned with the privileged basis vectors, with SRM plots \textit{Figs.}~\ref{Fig:Overcomplete1} and \ref{Fig:Overcomplete2} demonstrating this respectively. This shows that the SRM technique is also versatile to highly overcomplete privileged bases and that the distribution of activations continues to be affected by the choice of functional form for the activation functions --- developing the same or opposing anisotropies to the generalised $\tanh$'s anisotropies.

\newpage
\section{Generalised Tanh Activation Function}
\label{App:GeneralTanh}

    This paper uses modified and novel versions of the $\tanh$ function to form a functional class. They are modified such that an arbitrary basis of varying completeness can be used to construct a $\tanh$-like function. The $\tanh$-like function's operation is then basis dependent on this arbitrary basis, as the basis vectors show up explicitly in its multivariate form. Therefore, the explicit dependence, results in anisotropy being about these directions and hence, this arbitrary basis becomes the privileged basis.
    
    The motivation for this activation function is two-fold: it shows the versatility of the SRM method for arbitrary privileged bases, and the decoupling of the privileged bases from the standard bases directly shows how functional form choices induce representational alignment. However, these are not expected to be a practical activation function in wider applications, unless a specific basis privileging is necessitated. This section discusses its derivation.
    
    In deep learning, $\tanh$ is typically applied elementwise to decomposed elements of the standard basis. This multivariate function is shown in \textit{Eqn.}~\ref{Eqn:AnisotropicTanh}, defining $\sigma:\mathbb{R}^n\rightarrow\mathbb{R}^n$ in terms of the standard basis vectors $\hat{e}_i$ for $i = 1,\ldots,n$. 
    
    This multivariate representation differs from its usual (oversimplified) univariate form which obfuscates the basis privileging.
    \begin{equation}
        \sigma\left(\vec{x}\right) = \sum_{i=0}^{n-1}\tanh\left(\vec{x}\cdot\hat{e}_i\right)\hat{e}_i
        \label{Eqn:AnisotropicTanh}
    \end{equation}
    However, a non-standard alternative (orthonormal) basis could be constructed, $\hat{b}_i$, and the $\tanh$ could be applied along its decomposed elements. This is shown in \textit{Eqn.}~\ref{Eqn:NonStandardAnisotropicTanh}.
    \begin{equation}
        \sigma\left(\vec{x}\right) = \sum_{i=0}^{n-1}\tanh\left(\vec{x}\cdot\hat{b}_i\right)\hat{b}_i
        \label{Eqn:NonStandardAnisotropicTanh}
    \end{equation}
    Furthermore, this basis can be made overcomplete, complete or incomplete by varying the number of basis vectors. So instead of having $n$ orthonormal basis vectors for a $\mathbb{R}^n$ space, $m$ 
    unit-vectors can be utilised for the same $\mathbb{R}^n$ space; in this work, it is important that they are Thompson bases such that
    these $m$ vectors are distributed evenly using a modified Thompson problem, discussed in \textit{App.}~\ref{App:ThompsonBasis}. For three dimensions, these bases sometimes form the corners of the platonic solids along with extra shapes such as triangular dipyramids.
    
    To prevent undesirable interference, only (Thompson) basis vectors with a positive dot-product with the input contribute. This is demonstrated in \textit{Eqn.}~\ref{Eqn:GeneralisedTanhUnnormalised}, defining $\sigma:\mathbb{R}^n\rightarrow\mathbb{R}^n$ in terms of the $m$ Thompson basis vectors. The use of a Thompson basis doubles the number of basis vectors required to produce a rotated elementwise $\tanh$ function. Therefore, for $m\leq n$, the basis is undercomplete, and the basis vectors form a $m-1$ dimensional simplex. For $m=n+1$, the basis is complete and forms an $n$ dimensional simplex. For $m>n+1$, the basis is overcomplete, with $m=2n$ reproducing the standard elementwise application but rotated arbitrarily.
    \begin{equation}
        \sigma\left(\vec{x}\right) = \sum_{i=0}^{m-1}\tanh\left(\max\left(0, \vec{x}\cdot\hat{b}_i\right)\right)\hat{b}_i
        \label{Eqn:GeneralisedTanhUnnormalised}
    \end{equation}
    In \textit{Eqn.}~\ref{Eqn:GeneralisedTanhUnnormalised}, if one were to observe the mapping of a one-dimension subspace corresponding to $\hat{b}_j$, it would be observed that the map no longer applies the $\tanh$ function when $m>2n$. Rather, it applies a summed series of scaled $\tanh$ functions. This is undesirable since it causes a discontinuous behaviour in the functional form for the class. Therefore, a correction 
    term is added to preserve this behaviour. For the correction, see \textit{Eqn.}~\ref{Eqn:GeneralisedTanhNormalised} below. It was chosen to be an addition of a function that takes the magnitude of the input. This choice prevents unexpected angular oddities in the mapping. It could be argued that the correction breaks the class structure for non-basis directions, but empirically, it was found to benefit performance by preserving $\tanh$ along basis directions. Additionally, the correction is only applied along positive dot-products as otherwise negative and positive direction contributions can be cancelled out. 
    \begin{equation}
        \sigma\left(\vec{x}\right) = \sum_{i=0}^{m-1}\tanh\left(\max\left(0, \vec{x}\cdot\hat{b}_i\right)\right)\hat{b}_i + \max\left(0, \hat{x}\cdot\hat{b}_i\right)N\left(\left\|\vec{x}\right\|\right)\hat{b}_i
        \label{Eqn:GeneralisedTanhNormalised}
    \end{equation}
    The anti-interference correction 
    term implicitly defined a quantity $N\left(\left\|\vec{x}\right\|\right)$ which can be derived in explicit form using the aforementioned one-dimensional slice with $\hat{b}_j$ but is valid for all $\hat{b}_j$. The derivation is shown below. 
    To start, \textit{Eqn.}~\ref{Eqn:GeneralisedTanhDerivation1} defines the desired equality:
    \begin{equation}
        \sigma\left(\alpha \hat{b}_j\right)\cdot\hat{b}_j:=\tanh\left(\alpha\right)
        \label{Eqn:GeneralisedTanhDerivation1}
    \end{equation}
    \textit{Eqn.}~\ref{Eqn:GeneralisedTanhDerivation2} is produced by substituting in the function $\sigma$ into \textit{Eqn.}~\ref{Eqn:GeneralisedTanhDerivation1}.
    \begin{equation}
        \sum_{i=0}^{m-1}\tanh\left(\max\left(0, \alpha\hat{b}_j\cdot\hat{b}_i\right)\right)\hat{b}_i\cdot\hat{b}_j + \max\left(0, \hat{b}_j\cdot\hat{b}_i\right)N\left(\alpha\right)\hat{b}_i\cdot\hat{b}_j := \tanh\left(\alpha\right)
        \label{Eqn:GeneralisedTanhDerivation2}
    \end{equation}
    Rearranging this last equation to isolate $N\left(\alpha\right)$ yields \textit{Eqn.}~\ref{Eqn:GeneralisedTanhDerivation3}.
    \begin{equation}
        N\left(\alpha\right) = \frac{\tanh\left(\alpha\right)-\sum_{i=0}^{m-1}\tanh\left(\max\left(0, \alpha\hat{b}_j\cdot\hat{b}_i\right)\right)\hat{b}_i\cdot\hat{b}_j}{\sum_{i=0}^{m-1}\max\left(0, \hat{b}_j\cdot\hat{b}_i\right)\hat{b}_i\cdot\hat{b}_j}
        \label{Eqn:GeneralisedTanhDerivation3}
    \end{equation}
    This can then be simplified to \textit{Eqn.}~\ref{Eqn:GeneralisedTanhDerivation4}.
    \begin{equation}
        N\left(\alpha\right) = -\frac{\sum_{i\neq j}\tanh\left(\alpha\max\left(0, \hat{b}_j\cdot\hat{b}_i\right)\right)\hat{b}_i\cdot\hat{b}_j}{\sum_{i=0}^{m-1}\max\left(0, \hat{b}_j\cdot\hat{b}_i\right)^2}
        \label{Eqn:GeneralisedTanhDerivation4}
    \end{equation}
    For exact Thompson bases, this function is constant for every $\hat{b}_j$; however, for approximate bases, an average over $j$ can be taken, as shown in \textit{Eqn.}~\ref{Eqn:GeneralisedTanhDerivation5}.
    \begin{equation}
        N\left(\alpha\right) = -\frac{1}{m}\sum_{j=0}^{m-1}\frac{\sum_{i\neq j}\tanh\left(\alpha\max\left(0, \hat{b}_j\cdot\hat{b}_i\right)\right)\hat{b}_i\cdot\hat{b}_j}{\sum_{i=0}^{m-1}\max\left(0, \hat{b}_j\cdot\hat{b}_i\right)^2}
        \label{Eqn:GeneralisedTanhDerivation5}
    \end{equation}
    This, with $\alpha = \left\|\vec{x}\right\|$, gives $N\left(\left\|\vec{x}\right\|\right)$ explicitly as the correction term. This can be substituted into \textit{Eqn.}~\ref{Eqn:GeneralisedTanhNormalised} to yield \textit{Eqn.}~\ref{Eqn:FinalGeneralTanh} below. It is this activation function functional class that is used across all results for various stated $m$ and $n$ values. For each particular $m$ and $n$ all valid Thompson bases are part of the functional class. In practice, this means the activation function's privileged basis $\{\hat{b}_j\}$ may be rotated arbitrarily. 
    \begin{align}
        \sigma\left(\vec{x}\right) 
        &= \sum_{i=0}^{m-1}\tanh\left(\max\left(0, \vec{x}\cdot\hat{b}_i\right)\right)\hat{b}_i 
        \nonumber
        \\
        &\hspace{15mm} -\max\left(0, \hat{x}\cdot\hat{b}_i\right)\frac{1}{m}\sum_{j=0}^{m-1}\frac{\sum_{i\neq j}\tanh\left(\left\|\vec{x}\right\|\max\left(0, \hat{b}_j\cdot\hat{b}_i\right)\right)\hat{b}_i\cdot\hat{b}_j}{\sum_{i=0}^{m-1}\max\left(0, \hat{b}_j\cdot\hat{b}_i\right)^2}\hat{b}_i
        \label{Eqn:FinalGeneralTanh}
    \end{align}
    It is not proposed that this activation function is in any way computationally or practically desirable; it is merely a tool to explore how activation functions can affect the privileging of a basis. Upcoming work will explore this function class' effect on performance for various $m$ and $n$ and crucially as $m\rightarrow \infty$. An alternative formulation could also be used as shown in \textit{Eqn.}~\ref{Eqn:MaxTanhForm}, which limits to an exciting new class of activation functions and networks to be termed as \textit{Isotropic Deep Learning} and is briefly discussed in \textit{App.}~\ref{App:Isotropic}
    \begin{equation}
        \sigma\left(\vec{x}\right) = \frac{\tanh\left(\left\|\vec{x}\right\|\right)\hat{x}}{\underset{{\hat{b}_i}}{\max}\left({\hat{b}_i\cdot\hat{x}}\right)}
        \label{Eqn:MaxTanhForm}
    \end{equation}

\newpage
\section{Producing a Thompson Basis} 
\label{App:ThompsonBasis}

    The Thompson basis is an attempt to (approximately) evenly distribute $m$ vectors in $\mathbb{R}^n$. Alternative methods were considered, such as Fibonacci lattices, though the approximation provided by the following method was better in terms of its distribution. An approximation is necessary as only certain values for $m$ produce exact vector arrangements in $\mathbb{R}^n$. This approach allows generalisation to other $m$ values where arrangement may not be known or be possible.

    A variety of Thompson-like bases generation methods are possible \citep{Tammes1930, Claxton1966, Erber1991, Altschuler1994}. To generate the bases in these experiments, PyTorch's gradient descent algorithm was used on the energy function shown in \textit{Eqn.}~\ref{Eqn:ThompsonEnergyFunc}, which is written using Einstein summation convention. The $m$ basis unit-vectors $\hat{b}_i\in\mathbb{R}^n$ are stacked row-wise into matrix $\mathbf{V}\in\mathbb{R}^{n\times m}$ and initialised normally. The $m$-by-$m$ identity matrix is denoted $\mathbf{I}$ whilst a $m$-by-$m$ matrix of all elements equal to one is denoted $\mathbf{1}$. The basis vectors $\hat{b}_i$, which forms the rows of $\mathbf{V}$, are constrained to unit-norm throughout training.
    \begin{equation}
        E = \mathbf{D}_{ij}\mathbf{V}_{ki}\mathbf{V}_{kj}\left(\mathbf{1}_{ij}-\mathbf{I}_{ij}\right)_{ij}
        \label{Eqn:ThompsonEnergyFunc}
    \end{equation}
    Matrix $\mathbf{D}$ is an inverse-pairwise-distance matrix, found to be empirically necessary to avoid cancellations between opposing directions, given elementwise by \textit{Eqn.}~\ref{Eqn:DistanceFunction}. If a divide-by-zero occurs, the value of that index is set to zero.
    \begin{equation}
        \mathbb{R}^{m\times m}\ni\mathbf{D}_{ij} =\left\{\begin{matrix} \frac{1}{\left\|\hat{b}_i-\hat{b}_j\right\|_2^2} &:&  i \neq j \\ 0 &:& i = j \\\end{matrix}\right. 
        \label{Eqn:DistanceFunction}
    \end{equation}
    This differs substantially from Thompson's electrostatic repulsion implementation, as it minimises pairwise similarity. This was found to be empirically advantageous when using gradient descent and provided a good and fast approximation of a Thompson Basis for the experiments.

\newpage
\section{Autoencoder Model Architectures}
\label{App:Architectures}
       
\begin{figure}[htb]
       \begin{center}
       \includegraphics[width=0.75\textwidth]{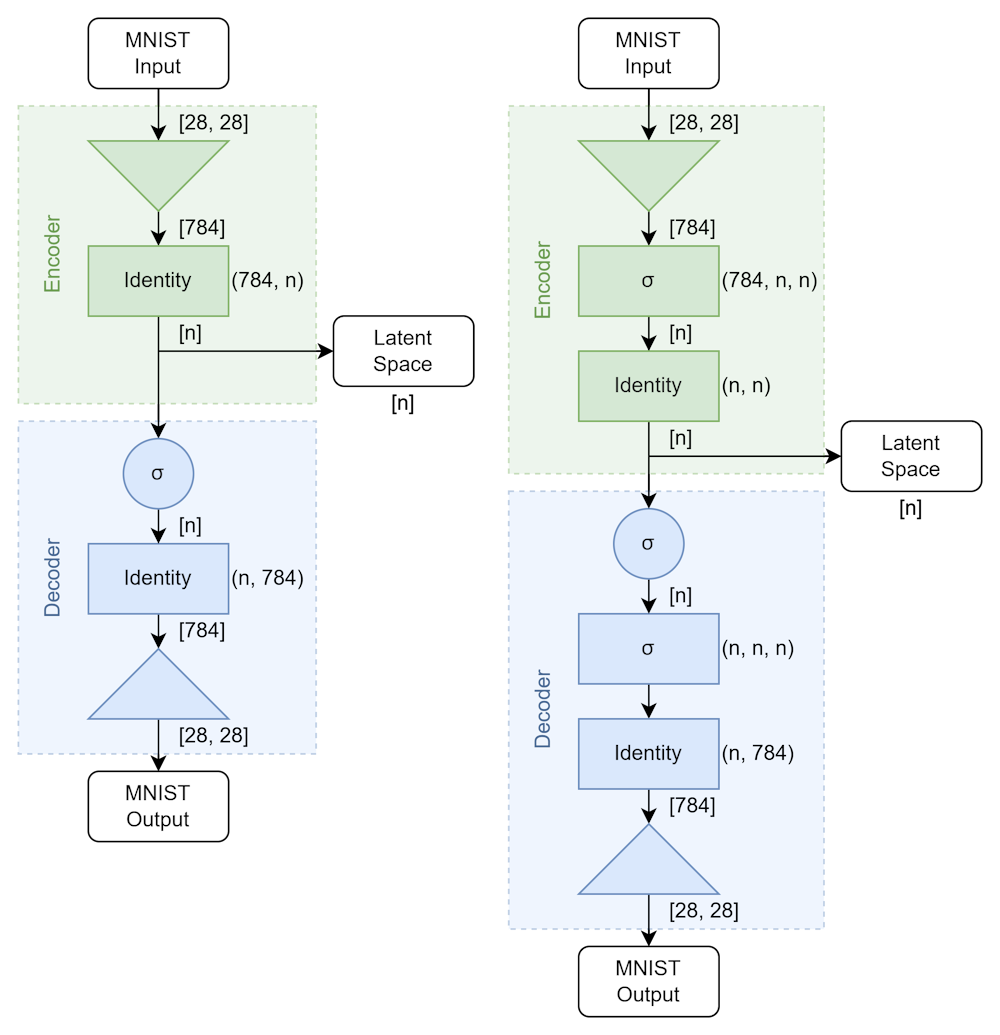}
       \end{center}
       \caption{Shows the autoencoding models used for the reconstruction of MNIST samples. The left plot shows the `small' model, whilst the right plot shows the `large' model. Both use linear layers and generalised tanh activation function $\sigma$. The architectures are displayed using the neural notation convention described in \textit{App.}~\ref{App:NeuralNotation}.}
       \label{Fig:MNISTarchitectures}
    \end{figure}

    The figures below show all the architectures of the networks used in this paper. They are illustrated using the neural notation convention described in \textit{App.}~\ref{App:NeuralNotation}. \textit{Figure}~\ref{Fig:MNISTarchitectures} shows the small and large architectures for the MNIST autoencoders. The value $n$ is the neuron number of the hidden layer, which will be listed per result alongside the number of privileged basis vector directions $m$. For training, a batch size of $24$, a learning rate of $0.08$ and $100$ epochs were used to standardise across all networks. These values are largely arbitrary but offered good empirical performance on the reconstruction --- though no algorithmic fine-tuning of these hyperparameters was done. It is the `small MNIST' model depicted in \textit{Fig.}~\ref{Fig:MNISTarchitectures}, which is presented in the primary results of \textit{Sec.}~\ref{Sec:Results}. This particular architecture was chosen for the primary result since it provided a good representation of the overall results whilst also being the simplest and, therefore, interpretable model. It also has no prior activation function before the latent space --- as this could have reshaped the distribution more complexly, as observed in \textit{Sec.}~\ref{Sec:Elementwise}. The output latent space depicted in each figure is the resultant data on which SRM was computed in all cases. 

    The models are in four varieties: \textit{small} or \textit{large} and \textit{MNIST} or \textit{CIFAR}. Each consists of an `encoder' and `decoder', from which the activations of the latent layer will be analysed. 

\begin{figure}[htb]
       \begin{center}
       \includegraphics[width=0.75\textwidth]{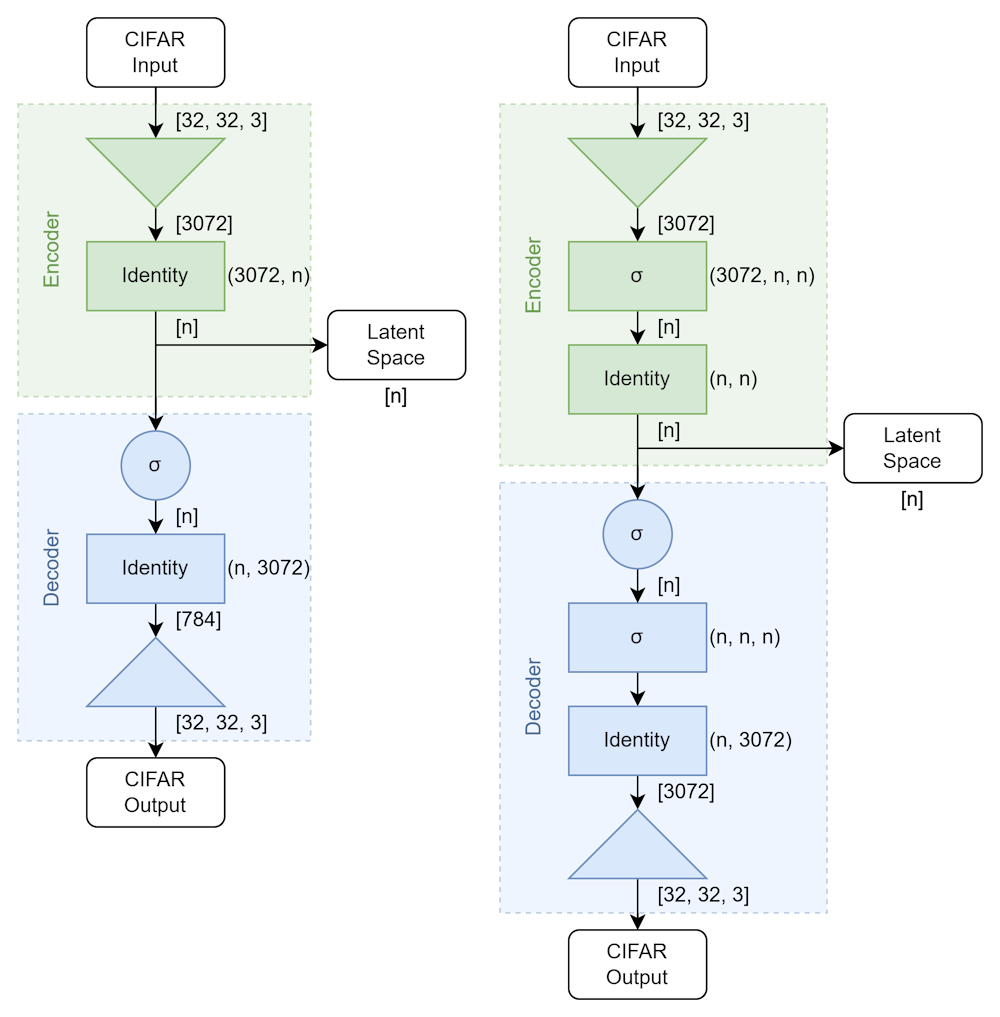}
       \end{center}
       \caption{Shows the autoencoding models used for the reconstruction of CIFAR samples. The left plot shows the `small' model, whilst the right plot shows the `large' model.  Both use linear layers and generalised tanh activation function $\sigma$. The architectures are displayed using the neural notation convention described in \textit{App.}~\ref{App:NeuralNotation}.}
       \label{Fig:CIFARarchitectures}
    \end{figure}

    For consistency of interpretation, the autoencoder architecture for the CIFAR reconstruction is similar to MNIST. They are shown in \textit{Fig.}~\ref{Fig:CIFARarchitectures}
    Extra demonstrations of the SRM technique on these extra network architectures are shown in \textit{App.}~\ref{App:ExtraTests}.
    
\newpage
\section{Ratio of volumes of an n-segment to n-ball}
\label{App:nBall}

    Assuming a uniform, infinitely sampled, embedded dataset, the expectation value $\mathbb{E}_\theta\left[f_{\text{SRM}}\right]$, is the ratio of the volumes of an $n$-segment to an $n$-ball, as given in \textit{Eqn.}~\ref{Eqn:nball} \citep[cf.][]{stackexchangeVolumeCone}.
    
    \begin{equation}
      \mathbb{E}_\theta\left[f_{\text{SRM}}\right]= \frac{\lambda^{n-1}\mathcal{B}^{n-1}_1}{\lambda^{n}\mathcal{B}^{n}_1}\left(\frac{2}{n}\sin^{n-1}\left(\phi\right)\cos\left(\phi\right)+B\left(\frac{1}{2},\frac{n+1}{2}\right)-B_{\cos^2\left(\phi\right)}\left(\frac{1}{2},\frac{n+1}{2}\right)\right)
      \label{Eqn:nball}
    \end{equation}
    
    With $B$ being the beta function, $B_{\cos^2\left(\phi\right)}$ the unnormalised incomplete beta function, $\mathcal{B}^{n}_1$ the volume of the unit-n-ball and $\lambda_n$ being the $n$-lebesgue measure.

\newpage
\section{Isotropic Approaches to Deep Learning}
\label{App:Isotropic}

    Many of the most commonly used functions in deep learning are basis-dependent to a particular basis (often the standard basis). This is not clear in many notations, which suppress this dependence by only writing univariate forms of the function. This oversimplification of the functions obfuscates the privileging of a basis in a deep learning model, which is likely unintentional by most developers. Furthermore, there are some functional forms that privilege opposing bases, namely dropout.

    In this paper, many model functional form choices were made to be isotropic. This prevented competing privileged bases from complicating the analysis, isolating activation functions as the sole cause for anisotropy. The functional form choices are reasoned below.

    Many formulations of gradient descent, including nearly all adaptive methods, privilege the standard basis in their formulation. For adaptive optimisers, this is usually due to a diagonal approximation of the Hessian allowing for $\mathcal{O}\left(n\right)$ time computation in the number of parameters, as opposed to the (isotropic) Newton method, which is $\mathcal{O}\left(n^2\right)$. Therefore, only (minibatch) standard gradient descent or momentum variations were feasible and permittable for isolating anisotropies to activation functions. Therefore, momentum gradient descent was used with a learning rate of $0.08$ and a momentum factor of $0.9$.

    Any \textit{standard} normal initialiser is isotropic due to the standard multivariate normal's rotational symmetry. This requires mean of $\vec{\mu}=\vec{0}$ and covariance of $\Sigma=\sigma \mathbf{I}_n$. Xavier-normal was simply chosen for its particular covariance matrix. Orthogonal initialisers are also isotropic; however, they may interact differently with the various completeness of the bases, and they may particularly favour elementwise $m=2n$ bases. Hence, it was not used. 

    To simplify the models, no regularisation or normalisation was used, though isotropic forms such as L2 and Zero-phase component analysis (ZCA) \citep{Bell1996} can be used. ZCA is effectively an unrotated form of principal component analysis.

    Finally, it was important that the task was reconstruction when isolating activation functions. Humans choose the final layer of classifiers to be human-interpretable. In practice, this typically means a one-hot basis, which makes the goal of training the network the production of an anisotropic function of the data. Using reconstruction prevents this privileging of a human-interpretable basis. Interaction, such as between the activation function's privileging of a basis against opposing output layer privileging, may explain the neural collapse phenomenon's relation to classification networks. Despite this, the data may still privilege a particular completeness of basis due to its hypercubic bounding: $\left[0, 1\right]^{28\times 28}$ or $\left[0, 1\right]^{32\times 32\times3}$ for MNIST or CIFAR respectively. However, this is inseparable from the dataset and similar in all datasets, so it is unavoidable unless using a toy dataset, such as reconstructing random normal vectors. It was felt that testing on the standard MNIST and CIFAR datasets would provide more interpretability and utility to the reader. The MNIST and CIFAR datasets were linearly rescaled to $\left[-1, 1\right]^{28\times 28}$ or $\left[-1, 1\right]^{32\times 32\times3}$ respectively for all reconstruction training, testing and analyses. This was to approximately centre the distributions at zero. 
    
    Overall, these isotropic functional form choices are essential if one wishes to determine a definitive privileged basis. This was required to establish the basic efficacy of SRM. Despite this, the isotropic functional form choices may be relaxed, and using SRM, a hierarchy could be constructed for which functions influence the privileging of the basis the most or even detect the presence of hybridized privileged bases perhaps present for phenomena like neural collapse. Isotropic approaches to deep learning, including isotropic-$\tanh$, are the primary topic of the author's PhD, so will be explored further in future work. 
\newpage
\section{Neural Architecture Notation Convention \label{App:NeuralNotation}}

    This section summarises the diagrammatic system used in the production of \textit{App.}~\ref{App:Architectures} figures. This system intends to make available and publically editable a standardised and centralised manner of depicting neural network architectures across papers to ease interpretation for the reader. The system is centralised on a GitHub page (\url{https://github.com/GeorgeBird1/Diagramatic-Neural-Networks}) which can be edited by the community.

\begin{figure}[htb]
        \begin{center}
        \includegraphics[width=0.98\textwidth]{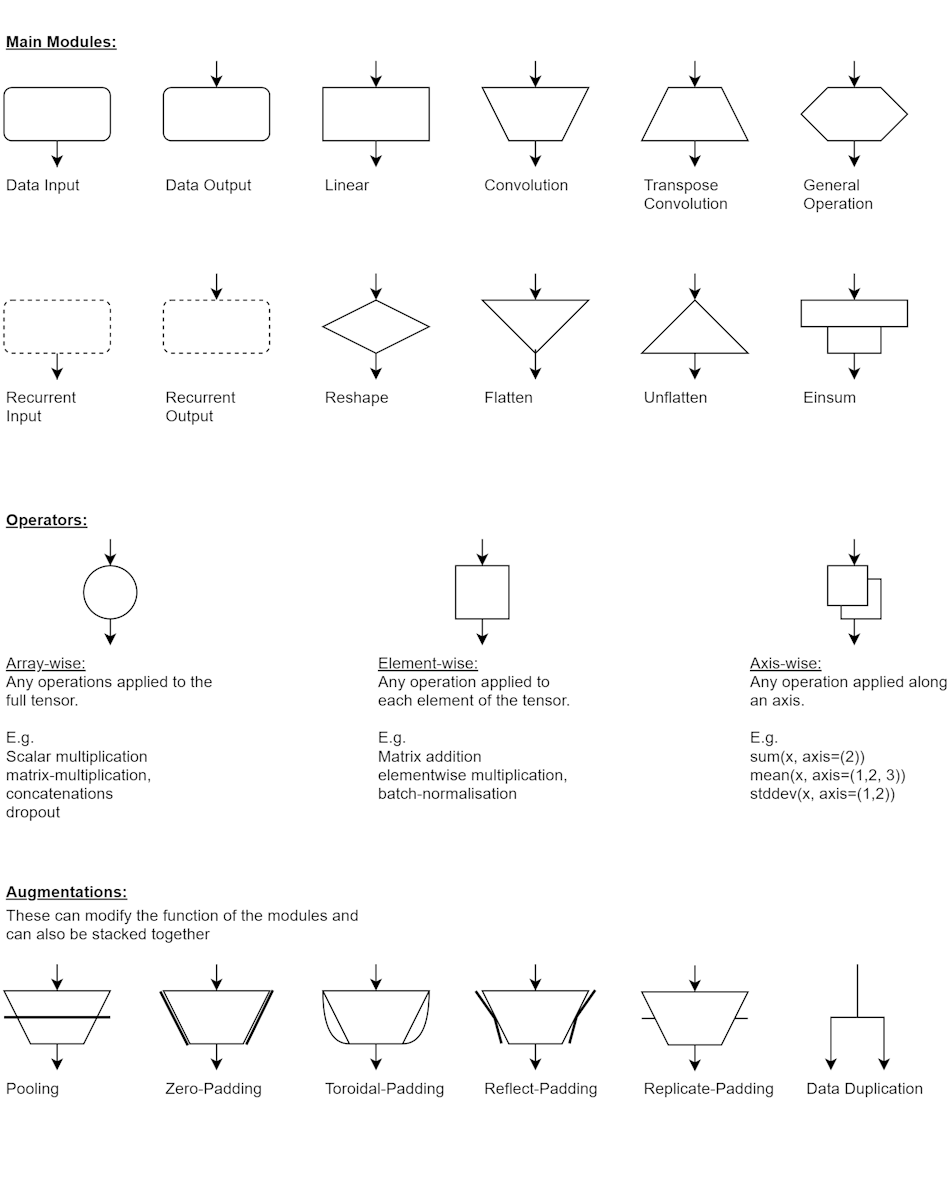}
        \end{center}
        \caption{Shows the basic modules which can be used in the system. These are common architectural blocks which appear in many models. The augmentation section is typically used for convolutional blocks, indicating how padding should be applied.}
       \label{Fig:DiagrammaticOne}
    \end{figure}

\begin{figure}[htb]
        \begin{center}
        \includegraphics[width=0.98\textwidth]{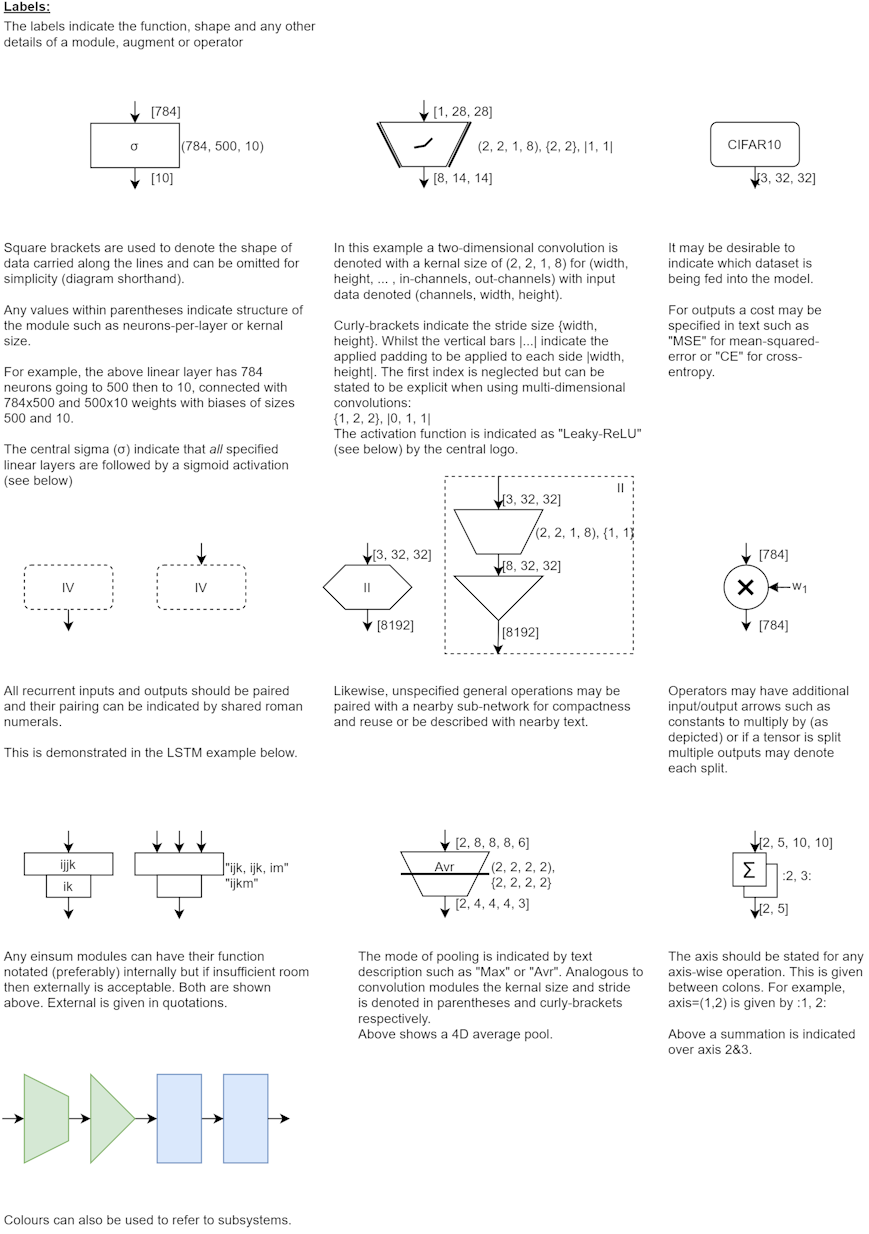}
        \end{center}
        \caption{Shows how the modules from \textit{Fig.}~\ref{Fig:DiagrammaticTwo} can have extra information added to detail its specific implementation in a model.}
        \label{Fig:DiagrammaticTwo}
    \end{figure}

    The system is broken down into two figures: \textit{Figs.}~\ref{Fig:DiagrammaticOne} and \ref{Fig:DiagrammaticTwo}.
    
    \newpage
    
\end{document}